\documentclass[11pt]{article}

\newcommand{\res}[1]{\ensuremath{R_#1}}

\newcommand{\dc}[1]{\textsc{de-colloquialisation}}

\usepackage{booktabs}
\usepackage{multirow}
\usepackage{adjustbox}
\usepackage{xcolor}
\newcommand{\chg}[1]{\textbf{#1}}
\usepackage{tabularx}
\usepackage{xurl}

\usepackage[table]{xcolor}
\definecolor{BestRowGray}{gray}{0.93}
\newcommand{\bestrow}{\rowcolor{BestRowGray}}
\newcommand{\best}[1]{\textbf{#1}}

\usepackage{amsmath, amssymb}
\usepackage{adjustbox}

\usepackage{wrapfig}

\usepackage[preprint]{acl}
\usepackage{times}
\usepackage{latexsym}

\usepackage[T1]{fontenc}
\usepackage[utf8]{inputenc}
\usepackage{microtype}
\usepackage{inconsolata}
\usepackage{graphicx}

\usepackage{float}

\usepackage{paralist}

\title{BiCon-Gate: Consistency-Gated De-colloquialisation for Dialogue Fact-Checking}

\author{Hyunkyung Park \and
  Arkaitz Zubiaga \\
  Queen Mary University of London \\
  \texttt{\{hyunkyung.park, a.zubiaga\}}@qmul.ac.uk \\}

\begin{document}
\maketitle

\begin{abstract}
Automated fact-checking in dialogue involves multi-turn conversations where colloquial language is frequent yet understudied. To address this gap, we propose a conservative rewrite candidate for each response claim via staged de-colloquialisation, combining lightweight surface normalisation with scoped in-claim coreference resolution. We then introduce BiCon-Gate, a semantics-aware consistency gate that selects the rewrite candidate only when it is semantically supported by the dialogue context, otherwise falling back to the original claim.
This gated selection stabilises downstream fact-checking and yields gains in both evidence retrieval and fact verification. On the DialFact benchmark, our approach improves retrieval and verification---with particularly strong gains on SUPPORTS---and outperforms competitive baselines, including a decoder-based one-shot LLM rewrite that attempts to perform all de-colloquialisation steps in a single pass.
\end{abstract}

\section{Introduction}

Automated fact-checking determines whether a claim is supported, refuted, or there is not enough information (NEI) to verify it by grounding the claim in evidence from a reliable knowledge source such as Wikipedia \cite{thorne-vlachos-2018-automated, thorne-etal-2018-fever}. Most systems follow a retrieve--verify pipeline: evidence retrieval (Information Retrieval; IR) followed by fact verification (FV).

\begin{figure}[t]
    \centering
    \includegraphics[width=\linewidth,trim=2mm 2mm 2mm 2mm,clip]{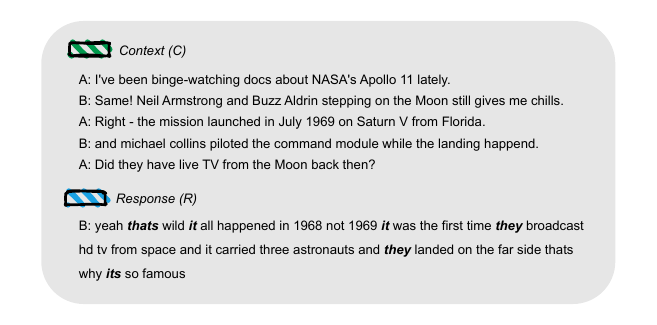}
    \caption{Example multi-turn dialogue illustrating a context-dependent response claim about the \textit{Apollo 11 mission}. In-claim pronouns in the response refer to antecedents introduced in earlier turns.}
    \label{fig:dialogue}
\end{figure}

Dialogue fact-checking extends this setting to multi-turn conversations, where a response claim depends on preceding context $C$ \cite{kim-etal-2021-robust}. In dialogue, colloquial phenomena---contractions, missing punctuation, inconsistent casing, ellipsis, and pronominal references---can blur claim boundaries and entity mentions (Figure~\ref{fig:dialogue}), making both retrieval and verification brittle \cite{kim-etal-2021-robust, chamoun-etal-2023-automated}.

A common remedy is to rewrite conversational inputs to be more self-contained (e.g., normalisation, de-contextualisation, or incomplete-utterance rewriting), which can improve recall on noisy queries \cite{sundriyal-etal-2023-chaos, li-etal-2023-incomplete, deng-etal-2024-document, guo-etal-2024-context, cao-etal-2024-incomplete}.

This creates a rewrite-induced IR--FV trade-off: making a claim more explicit can increase retrieval recall, while even small semantic drift (e.g., resolving a pronoun to the wrong entity) can mislead verification and degrade end-to-end (E2E) performance \cite{gupta-etal-2022-dialfact, chamoun-etal-2023-automated}. We focus on in-claim pronominal anaphora because it is frequent in dialogue claims and can often be addressed via minimal span substitution rather than broad paraphrasing; in DialFact, 42.1\% of claims contain at least one in-claim pronoun (\S\ref{sec:data}).

We ask whether we can improve retrieval and verification with minimal, meaning-preserving edits to the claim surface. To this end, we first construct a conservative rewrite candidate via staged de-colloquialisation: de-contraction, punctuation restoration, true-casing, and scoped in-claim pronoun resolution. We then propose \textbf{BiCon-Gate}, a bidirectional natural language inference (NLI)-based consistency gate that routes each instance to the rewrite candidate only when it is semantically supported by the dialogue context; otherwise, it falls back to the original claim \cite{zhang-etal-2023-relevance}.

Our contributions are three-fold:

\begin{compactitem}
 \item Controlling the IR--FV trade-off. We use a semantics-aware gate as a conservative control mechanism that selectively accepts rewrites to mitigate verification harm from semantic drift.
 \item Isolating where rewriting helps or hurts. We disentangle retrieval and verification effects via IR-only, FV-only (with gold evidence), and E2E evaluations.
 \item Demonstrating drift in one-shot LLM rewriting. We compare against a decoder-only, single-prompt rewrite baseline and show that aggressive one-shot rewrites can drift and hurt verification, whereas scoped edits paired with gating yield more reliable gains.
\end{compactitem}
To our knowledge, this is the first work that uses bidirectional NLI-style consistency signals as an instance-wise router for conservative dialogue-claim rewriting.

To structure our evaluation, we test three hypotheses. These hypotheses are stated as mechanism-driven expectations based on how the claim surface is consumed by both stages of the pipeline: IR uses it as a query, while FV uses it as the verifier's hypothesis. In particular, \res{1}--\res{3} make lightweight surface-form edits designed to preserve propositional content, whereas \res{4} makes explicit reference substitutions that can change downstream behaviour, motivating H3's test of whether semantic routing improves E2E robustness.

We denote the cumulative pipeline outputs as \res{1}--\res{4}, and \res{5} as the one-shot decoder rewrite (\S\ref{sec:method}). (H1) Lightweight surface normalisation (\res{1}--\res{3}) is largely neutral for both retrieval and verification. (H2) Resolving in-claim pronouns (\res{4}) can improve fact verification, and BiCon-Gate increases robustness by falling back when a rewrite is not semantically supported. (H3) In end-to-end fact-checking, semantic routing mitigates the tension between retrieval gains and verification robustness.

\begin{figure*}[t]
    \centering
    \includegraphics[width=\textwidth]{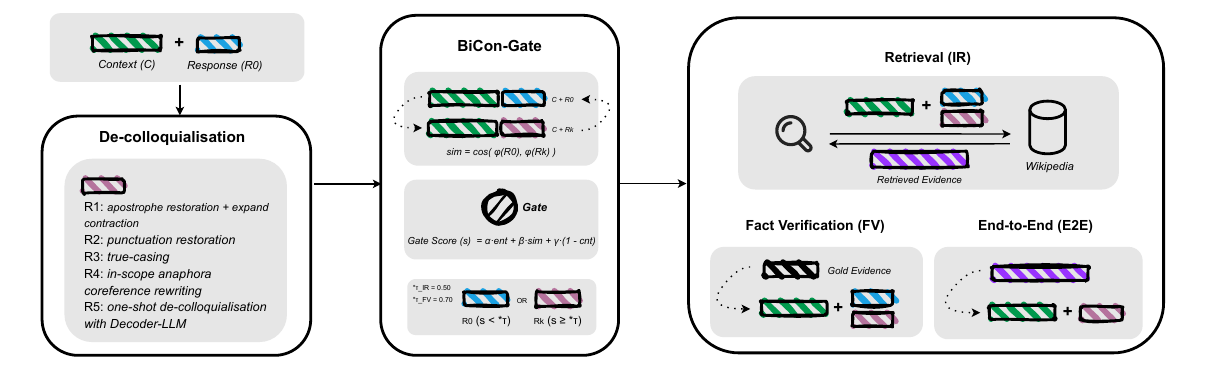}
    \caption{Overview of the staged de-colloquialisation pipeline (\res{1}-\res{5}), BiCon-Gate routing, and evaluation protocols (IR-only, FV-only, and E2E). BiCon-Gate combines bidirectional NLI entailment/contradiction signals (\textit{ent}/\textit{cnt}) with embedding cosine similarity (\textit{sim}) and accepts a rewrite only when the gate score exceeds a task-specific threshold ($\tau$); otherwise it falls back to \res{0}.}
    \label{fig:system-overview}
\end{figure*}

\section{Related Work}
\subsection{Colloquial noise in retrieval and verification}

Informal dialogue phenomena (e.g., contractions, ellipsis, and pronominal references) can blur span boundaries and entity mentions, making retrieval and verification brittle. In dialogue fact-checking, making a claim more explicit can improve evidence retrieval, but incorrect rewrites (e.g., wrong antecedent substitutions) can introduce semantic drift that harms downstream verification. This yields an IR--FV trade-off: aggressive de-colloquialisation may raise recall, while conservative, semantics-preserving rewriting is needed to maintain verification accuracy.

On the retrieval side, \citet{kim-etal-2021-robust} show that converting formal FEVER~\cite{thorne-etal-2018-fever} claims into colloquial variants led to a large drop in document recall (90.00\% $\rightarrow$ 72.20\%), highlighting how sensitive rankers are to informal phrasing. In dialogue, \citet{gupta-etal-2022-dialfact} similarly identify colloquiality---especially pronouns and underspecified references---as an obstacle to fact-checking.

A complementary line of work makes inputs more self-contained before retrieval. \citet{sundriyal-etal-2023-chaos} show that normalisation and de-contextualisation improve retrieval on noisy social media text, and de-contextualising pronoun-heavy claims improves evidence finding in multi-document settings~\cite{deng-etal-2024-document}. Related approaches rewrite incomplete utterances to mitigate ellipsis-driven failures in dialogue~\cite{li-etal-2023-incomplete, guo-etal-2024-context, cao-etal-2024-incomplete}. However, in fact-checking the rewritten claim also serves as the verifier's hypothesis; thus, retrieval-oriented rewrites can hurt E2E performance if they drift semantically. This motivates instance-wise safeguards that preserve meaning while still improving retrieval robustness.

Beyond rewriting, a related line of work improves fact-checking by explicitly aligning retrieval with verification objectives. Feedback-based Evidence Retriever (FER) feeds verifier signals back into the retriever, aligning document selection with end-task performance \cite{zhang-etal-2023-relevance}. Our work follows a similar principle---using downstream-oriented signals---but applies it to deciding when to trust a rewrite, rather than which documents to retrieve.

\subsection{Pronominal Coreference in Conversation}
Recent mention-based coreference resolution systems such as Maverick~\cite{martinelli-etal-2024-maverick} achieve strong accuracy with efficient span-scoring, but they are trained to link textual mentions and are not designed for discourse-level references or informal dialogue. We focus specifically on in-claim pronominal anaphora because the claim text is used directly as (i) the retrieval query and (ii) the verifier's hypothesis; unresolved pronouns therefore degrade both stages. Moreover, in-claim pronouns can often be resolved from the preceding dialogue without rewriting the entire conversation, making them a targeted and controllable intervention.

LLMs demonstrate impressive referential reasoning, but converting that ability into stable gains on standard coreference benchmarks---especially for pronominal mentions in long contexts---remains challenging~\cite{gan-etal-2024-assessing, manikantan-etal-2025-identifyme}. Recent analyses suggest that remaining errors concentrate on pronominal mentions and overlapping mention structures in long contexts, making antecedent substitutions brittle even for strong LLMs.

In dialogue fact-checking, \citet{chamoun-etal-2023-automated} find that naively substituting in-claim pronouns improves document recall (56.85\% $\rightarrow$ 67.00\%) but slightly harms sentence-level evidence selection (54.19\% $\rightarrow$ 53.86\%) and fact-verification accuracy (44.06\% $\rightarrow$ 42.71\%). These findings suggest that imperfect coreference rewrites can propagate errors downstream. Their error analysis attributes this drop to incorrect coreference links that steer evidence selection toward similar sentences in the wrong document and can even change a claim's label with respect to the gold evidence, motivating our instance-wise semantic gate.

Prior work typically studies individual rewrite phenomena in isolation and reports only end-to-end outcomes, which conflates retrieval and verification effects under a fixed backbone. It also rarely provides an explicit, conservative acceptance criterion to prevent semantic drift when applying coreference-based rewrites. We address these gaps by (i) measuring the component-wise impact of cumulative rewrites on the same instances using IR-only, FV-only, and E2E protocols, and (ii) introducing an instance-wise semantic gate that adopts a rewrite only when it is consistent with the original claim.

\subsection{NLI-based factual consistency and semantic gating}
\label{sec:gate}
A recurring theme in factual-consistency research is to compare a candidate text against a reference via NLI-style containment and semantic similarity. SummaC operationalises consistency by checking whether each summary sentence is entailed by, and not contradicted by, the source document \cite{laban2021summacrevisitingnlibasedmodels}. AlignScore similarly defines alignment as being fully supported by the reference without contradiction \cite{zha2023alignscoreevaluatingfactualconsistency}. MENLI further shows that NLI-based entailment and contradiction signals are more robust indicators of correctness than sentence-similarity scores alone \cite{chen2023menlirobustevaluationmetrics}.

We adopt this principle in a different setting: rather than using NLI scores purely for evaluation, we use them operationally to decide whether a rewrite should be applied for a given instance. This design directly targets the IR--FV trade-off described above by accepting rewrites only when they are semantically consistent with the original claim. Motivated by NLI-based factual-consistency work (e.g., SummaC, AlignScore, MENLI), BiCon-Gate uses bidirectional entailment/contradiction signals together with semantic similarity to conservatively accept rewrites and otherwise fall back to the original claim.

\section{Task \& Data}
\label{sec:task-and-data}

\subsection{Task Definition}

Given a multi-turn dialogue context $C$ and a response claim $R$, our system retrieves evidence from a Wikipedia snapshot aligned with DialFact (see \S~\ref{sec:ir-only} for details) and predicts a label in \{SUPPORTS, REFUTES, NEI\}. Because the claim surface is used both as the retrieval query and as the verifier hypothesis, rewriting can have opposite effects on IR and FV (e.g., higher recall but increased semantic drift). We denote the original claim $R$ as \res{0} and the variants produced by our de-colloquialisation pipeline as \res{1}--\res{5}.

To isolate rewriting effects, we keep the retriever and verifier fixed and vary only the claim surface presented to them. We report results under three evaluation protocols: (i) IR-only, where the selected surface is used as the retrieval query; (ii) FV-only, where the verifier receives gold evidence and the selected surface is used as the hypothesis; and (iii) E2E, where both retrieval and verification consume the selected surface.

Figure~\ref{fig:system-overview} summarises our pipeline and where rewriting interacts with retrieval and verification. Starting from the original claim \res{0}, we generate progressively de-colloquialised variants \res{1}--\res{4} through lightweight surface normalisation (de-contraction, punctuation restoration, true-casing) followed by scoped in-claim coreference rewriting. We also construct an alternative one-shot LLM rewrite \res{5}. BiCon-Gate then routes between \res{0} and a candidate rewrite (mainly \res{4}; \res{5} in an ablation) using semantic consistency signals derived from bidirectional NLI and embedding similarity; implementation details are provided in \S~\ref{sec:method}.

\subsection{Data}
\label{sec:data}

We use the official validation and test splits of DialFact \cite{gupta-etal-2022-dialfact}, whose dialogues are categorised into factual or personal subsets (Table~\ref{tab:dialfact-stat}). DialFact features colloquial multi-turn contexts, context-dependent claims, and a large proportion of NEI cases. We focus on DialFact because it combines multi-turn dialogue with a subset of sentence-level gold evidence, enabling retrieval-focused evaluation while still supporting E2E fact verification.

Sentence-level gold evidence annotations are provided only for the factual subset; accordingly, we compute IR metrics on that subset, while FV-only and E2E metrics are reported on the full test split. As shown in Table~\ref{tab:dialfact-stat}, 39.7\% of validation claims and 44.3\% of test claims contain at least one in-claim pronoun (42.1\% overall). This prevalence motivates our focus on in-claim anaphora: unresolved pronominal references are common and directly affect both retrieval queries and verifier hypotheses.

\section{Methodology}
\label{sec:method}
Our methodology consists of a staged de-colloquialisation pipeline with a semantic consistency gate. We describe the rewriting steps (\res{1}-\res{5}) and then the consistency gate and evaluation protocols.

\subsection{De-colloquialisation}
\label{sec:decolloquialisation}
We adopt a cumulative pipeline that reduces colloquial noise and then resolves in-claim anaphora. Let \res{0} be the original response; \res{1}--\res{5} are derived as follows.

\paragraph{\textbf{\res{1} De-contraction.}}
We first restore missing apostrophes with conservative, regex-gated rules (e.g., \textit{im/ive/ill} $\rightarrow$ \textit{I'm/I've/I'll}), then expand contractions (e.g., \textit{it's} $\rightarrow$ \textit{it is}) while protecting dotted acronyms (e.g., Ph.D., U.S.), honorifics, and URLs via placeholders. This stabilises token boundaries and negation cues.

\paragraph{\textbf{\res{2} Turn-preserving punctuation restoration.}}
We apply a multilingual punctuation model \cite{vandeghinste2023fullstoppunctuationsegmentationpredictiondutch} to each turn and
insert only predicted commas and sentence-final marks, leaving existing tokens and punctuation intact. For non-question turns without sentence-final punctuation, we append a period. This improves sentence and NP boundaries for downstream resolution.

\paragraph{\textbf{\res{3} True-casing.}}
Using a BERT-based masked LM (\texttt{bert-base-cased} \cite{devlin2019bertpretrainingdeepbidirectional}), we true-case sentence onsets and proper names with a margin rule: for each alphabetic token, we compare the MLM log-probabilities of upper- vs.\ lower-initial variants and flip only when the margin exceeds a threshold. We keep spelling, whitespace, and punctuation intact, modifying only token-initial characters to aid coreference resolution without semantic change of the underlying text.

\paragraph{\textbf{\res{4} Scoped coreference rewriting (with gate).}}

\underline{Scope.} We target only \emph{in-scope} pronominal anaphora whose antecedents are present in $C$ (e.g., \emph{he/she/they/you}), excluding deictic (\emph{this/that}) and expletive \emph{it} (e.g., \emph{it is raining}).
\underline{Detect \& propose.} On \res{3} we detect pronominal anchors (POS patterns) and let Maverick \cite{martinelli-etal-2024-maverick} propose up to 10 candidate antecedent NPs from the true-cased context.
\underline{Select \& rewrite.} We rank up to 10 candidates with an instruction-tuned LLM (\texttt{Llama-3.1-8B-Instruct} \cite{grattafiori2024llama3herdmodels}) and substitute the selected antecedent span for the pronoun mention in \res{3} to obtain a candidate \res{4}.

Instruction-following LLMs have been shown to work effectively as controllable decision modules for claim matching in automated fact-checking, motivating our use of an instruction-tuned LLM as a lightweight \emph{selector} over a small candidate referent set \cite{pisarevskaya2025zeroshotfewshotlearninginstructionfollowing}. We then apply BiCon-Gate (\S\ref{sec:bicongate}) as an instance-wise router.

\paragraph{\textbf{\res{5} Decoder-based one-shot reformulation.}}
A single \texttt{Qwen2.5-14B-Instruct} \cite{hui2024qwen2} prompt attempts all editing steps (\res{1}-\res{4}) in one shot given $\{C{+}\res{0}\}$, producing \res{5}. We use \texttt{Qwen2.5-14B-Instruct} as a representative strong instruction-tuned decoder to instantiate a competitive one-shot baseline.

The rewrite is prompted as a constrained editing task (Appendix Table~\ref{tab:prompt}); the model is instructed to apply only surface normalisation and unambiguous pronoun substitution based on the provided context, without adding or changing meanings.

For ablation experiments (\S\ref{sec:r5-ablation}) we optionally apply the same semantic gate as a router between \res{5} and \res{0}: for each instance, if $s_i^{(5)} \ge \tau$ we use \res{5}, otherwise we fall back to \res{0}. Model identifiers for all third-party components used in our pipeline are listed in the Appendix Table~\ref{tab:model-identifiers}.

\subsection{Consistency Gate: BiCon-Gate}
\label{sec:bicongate}
For an instance $i$ with context $C_i$ and original response \res{{0,i}}, a rewrite \res{{k,i}} ($k\in\{4,5\}$) is scored by
\begin{align}
    e_i
        &= \min\Bigl\{
            p^{\text{ent}}\bigl(C_i{+}\res{{0,i}} \Rightarrow \res{{k,i}}\bigr),
            \nonumber\\[-0.2em]
        &\qquad\quad
            p^{\text{ent}}\bigl(\res{{k,i}} \Rightarrow C_i{+}\res{{0,i}}\bigr)
            \Bigr\}, \label{eq:gate-ent} \\[0.1em]
    c_i
        &= \max\Bigl\{
            p^{\text{ctr}}\bigl(C_i{+}\res{{0,i}} \Rightarrow \res{{k,i}}\bigr),
            \nonumber\\[-0.2em]
        &\qquad\quad
            p^{\text{ctr}}\bigl(\res{{k,i}} \Rightarrow C_i{+}\res{{0,i}}\bigr)
            \Bigr\}, \label{eq:gate-ctr} \\[0.1em]
    \mathrm{sim}_i
        &= \cos\bigl(\phi(\res{{0,i}}),\,\phi(\res{{k,i}})\bigr). \label{eq:gate-sim}
\end{align}
Here $p^{\text{ent}}(\cdot)$ and $p^{\text{ctr}}(\cdot)$ denote calibrated NLI probabilities for the entailment and contradiction classes, respectively, so that $e_i, c_i \in [0,1]$. $\phi$ is a sentence encoder and $\mathrm{sim}_i$ is the cosine similarity between the original and rewritten claim. We take the minimum bidirectional entailment to require mutual semantic containment, and the maximum bidirectional contradiction to penalise any directional inconsistency.

We use a three-way NLI model and score both directions by swapping the premise and hypothesis between $C_i{+}\res{{0,i}}$ and $\res{{k,i}}$; we take $p^{\text{ent}}$ and $p^{\text{ctr}}$ from the entailment and contradiction softmax probabilities.

We calibrate the NLI probabilities via temperature scaling. Concretely, we rescale NLI logits $\mathbf{z}$ as $\mathbf{z}/T$ before softmax and learn a single scalar $T$ on the DialFact validation split (used as a calibration set) by minimising NLL. We then apply the learned $T=4.96$ to all bidirectional NLI scores at test time~\cite{xie2024calibratinglanguagemodelsadaptive, guo2017calibrationmodernneuralnetworks}.

The gate score and decision are:
\begin{align}
    s_i^{(k)}
        &= \alpha\,e_i {+} \beta\,\mathrm{sim}_i {+} \gamma\,(1{-}c_i), \label{eq:gate-score} \\
    \mathrm{accept}_i^{(k)}
        &= \mathbb{I}\bigl[s_i^{(k)} \ge \tau\bigr], \label{eq:gate-accept}
\end{align}

with non-negative weights $\alpha,\beta,\gamma\!\ge\!0$ that sum to one ($\alpha{+}\beta{+}\gamma=1$) and a threshold $\tau \in [0,1]$. For \res{{k,i}}, if $\mathrm{accept}_i^{(k)}=1$ we keep \res{{k,i}}; otherwise we fall back to \res{{0,i}}.

\subsection{Protocols and metrics}

\label{sec:protocol}
We evaluate each claim surface \res{k} under three protocols that disentangle retrieval and verification. Unless stated otherwise, IR/E2E results focus on \res{0}--\res{4}, and \res{5} is reported only as an FV-only ablation (\S\ref{sec:r5-ablation}); metrics are listed in Table~\ref{tab:metrics}.

\paragraph{IR-only.} The query is $C{+}\res{k}$ and the retriever is fixed.
We report document-level Recall@K and nDCG@K at the operating depths used in our retrieval stack (see \S\ref{sec:ir-only}).

For gate tuning (\S\ref{sec:gate-parameters}) we additionally report micro Recall@K and $1-\mathrm{ZHR@K}$. $\mathrm{ZHR@K}$ denotes the zero-hit rate (the fraction of queries whose top-$K$ retrieved set contains no gold evidence).

\paragraph{FV-only.}
We isolate the effect of the claim surface \res{k} on fact verification by providing the verifier with gold evidence sentences as premises and using $C{+}\res{k}$ as the hypothesis. We report fact-verification accuracy, macro-F1, and classwise F1 for SUPPORTS/REFUTES/NEI.

\paragraph{End-to-End (E2E).} IR and FV both consume \res{k}.
Unlike FV-only, the verifier receives the top-1 passage retrieved by the IR component as evidence, so E2E reflects the combined effect of rewriting under retrieval noise. We report the same FV metrics as FV-only: accuracy, macro-F1, and classwise F1.

\section{Experiments}

Our goal is to isolate the effect of claim de-colloquialisation and semantic routing, rather than to optimise the underlying retriever or verifier. Therefore, across all settings we keep the IR and FV backbones fixed and vary only the claim surface (\res{0}--\res{4}) and whether BiCon-Gate routes to a rewrite candidate; we additionally include a decoder-based one-shot rewrite (\res{5}) as an ablation.
We evaluate on DialFact because it provides multi-turn dialogue contexts and evidence annotations that support IR-only, FV-only, and E2E protocols.

\subsection{Gate parameters}
\label{sec:gate-parameters}

Following \S\ref{sec:gate}, we fix the gate weights to $(\alpha,\beta,\gamma){=}(0.4,0.2,0.4)$ and tune task-specific thresholds on the validation split. We set $\alpha$ and $\gamma$ symmetrically to weight entailment and non-contradiction equally, and down-weight cosine similarity ($\beta$) as a secondary signal because similarity alone can be high even under subtle semantic drift.

To minimise hyperparameter tuning while keeping the gate interpretable, we keep these weights fixed across all experiments and tune only $\tau$. We leave weight learning or re-tuning under larger distribution shifts to future work.

\paragraph{Threshold for IR.}
\begin{figure}[t]
    \centering
    \includegraphics[width=\linewidth,clip]{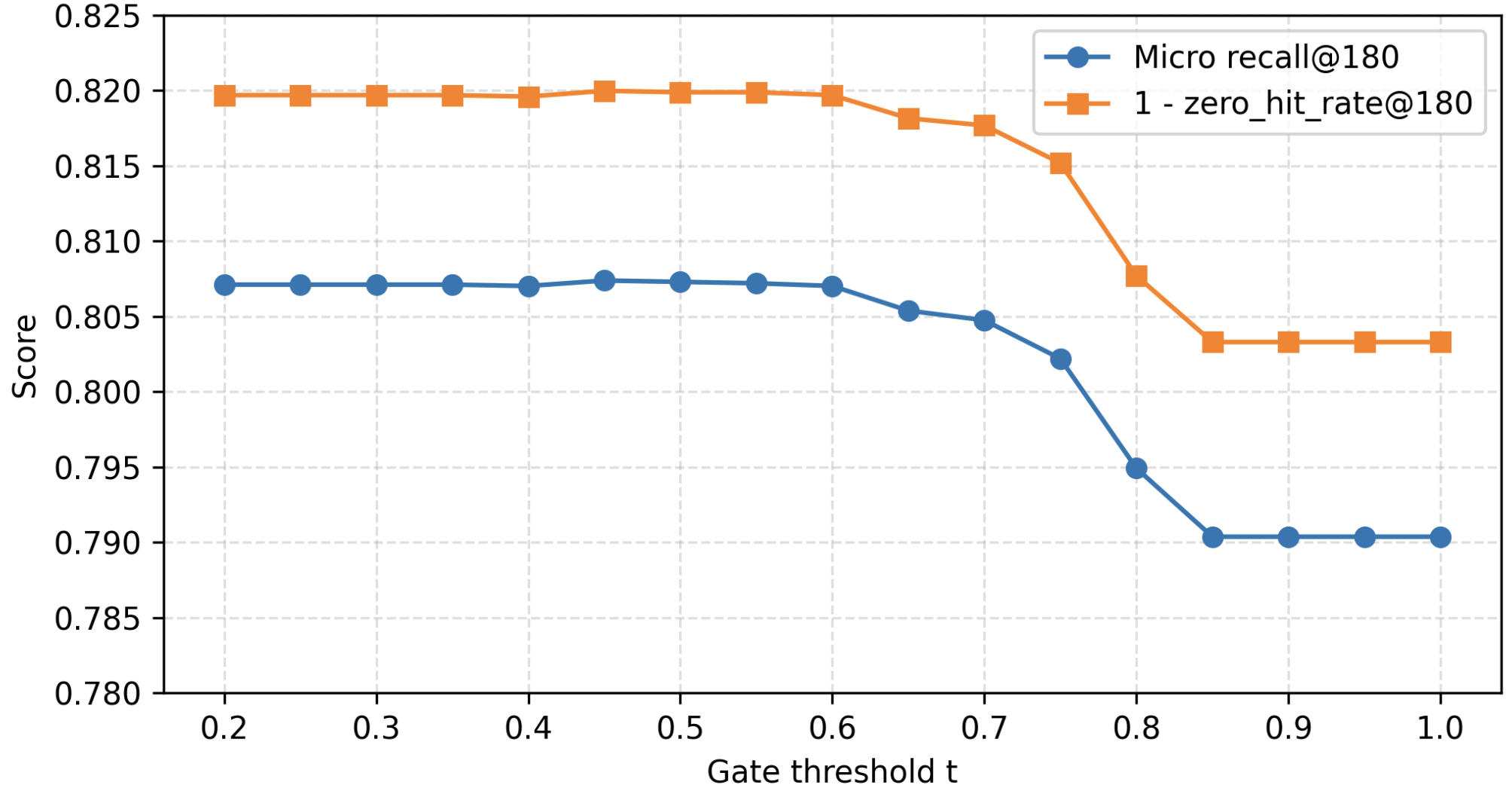}
    \caption{IR gate sweep on the validation split. We report BM25 micro-Recall@180 and $1{-}\mathrm{ZHR}@180$ as a function of the \res{4} gate threshold $\tau$ (where $\mathrm{ZHR}@180$ is the zero-hit rate). We use $\tau_{\text{IR}} = 0.50$ in the IR-only protocol.
    }
    \label{fig:valid_ir_metrics_t}
\end{figure}

For IR, we tune the threshold using BM25 retrieval metrics on the validation split. For each $\tau \in [0.20, 1.00]$ in increments of 0.05, we build a gated claim surface exactly as in \S\ref{sec:gate}, run BM25 with depth $K{=}180$, and compute micro-recall@180 and $1{-}\text{ZHR@180}$.

Both curves exhibit a broad plateau for $\tau \in [0.4,0.6]$ and start to degrade once $\tau>0.6$. We set the IR gate threshold to $\tau_{\text{IR}}{=}0.50$, which lies near the centre of this plateau and slightly maximises micro-recall@180 (Figure~\ref{fig:valid_ir_metrics_t}).

\paragraph{Threshold for FV.}
\begin{figure}[t]
    \centering
    \includegraphics[width=\linewidth,clip]{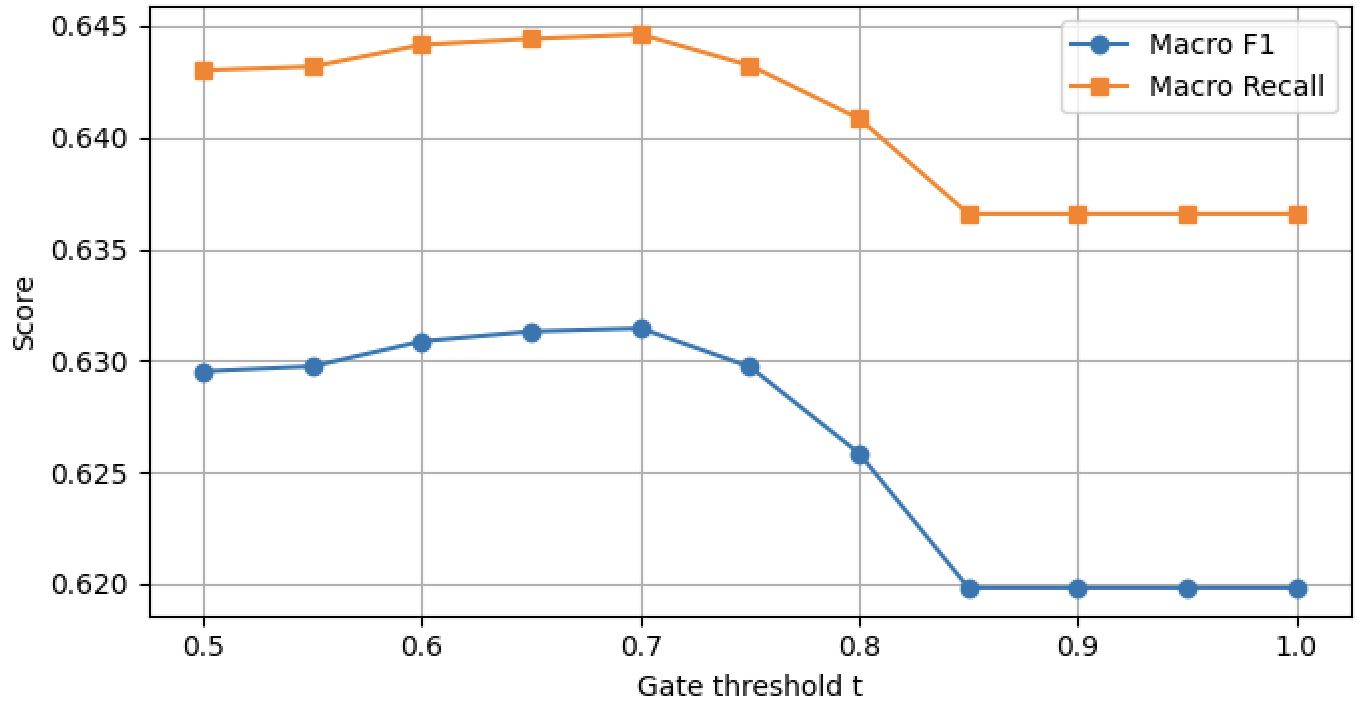}
    \caption{
        FV-only gate sweep on the validation split (gold evidence). Macro-F1 and macro-Recall as a function of the \res{4} gate threshold $\tau$; we use $\tau_{\text{FV}}=0.70$ in FV-only and E2E.
    }
    \label{fig:valid_f1_t_r4}
\end{figure}

For FV, we use gold evidence and sweep the gate threshold $\tau \in [0.20, 1.00]$ on the validation split. We use the \texttt{gateSweep} routine, which for each threshold $\tau$ constructs hypotheses with either \res{4} or \res{0} according to whether the BiCon-Gate score satisfies $s_i^{(4)} \ge \tau$, and runs the verifier once on the resulting premise--hypothesis pairs. As shown in Figure~\ref{fig:valid_f1_t_r4}, macro-F1 and macro-Recall both peak at $\tau\approx0.70$ on valid split; performance improves as the gate starts accepting high-scoring \res{4} rewrites, but once $\tau$ becomes too large we discard too many well-rewritten \res{4} and the curves drop back towards the \res{0} baseline.

\subsection{IR-only}
\label{sec:ir-only}
\begin{table}[t]
\centering
\footnotesize
\resizebox{\columnwidth}{!}{
    \setlength{\tabcolsep}{3pt}
    \begin{tabular}{lcccccc}
    \toprule
    & \multicolumn{2}{c}{BM25 (K=180)}
    & \multicolumn{2}{c}{E5 dense (K=10)}
    & \multicolumn{2}{c}{BGE-CE (K=1)} \\
    \cmidrule(lr){2-3} \cmidrule(lr){4-5} \cmidrule(lr){6-7}
    Claim & R@180 & nDCG@180 & R@10 & nDCG@10 & R@1 & nDCG@1 \\
    \midrule
    \res{0}
      & 73.89 & 48.37 & 70.47 & 61.98 & 53.51 & 57.01 \\
    \res{1}
      & 73.22 & 47.96 & 69.95 & 61.56 & 53.14 & 56.59 \\
    \res{2}
      & 73.22 & 47.96 & 69.94 & 61.60 & 53.18 & 56.64 \\
    \res{3}
      & 73.22 & 47.96 & 69.96 & 61.67 & 53.21 & 56.66 \\
    \res{4}
      & \textbf{76.54} & \textbf{50.94}
      & 73.18 & 64.45
      & 55.53 & 59.15 \\
    \textit{+Gated}
      & 76.53 & 50.91
      & \textbf{73.19} & \textbf{64.46 }
      & \textbf{55.58} & \textbf{59.19} \\
    \bottomrule
    \end{tabular}
}
\caption{IR-only retrieval results on the test (factual) split (macro Recall/nDCG, \%) for claim variants \res{0}--\res{4}. \res{4}+\textit{Gated} applies BiCon-Gate with $\tau_{\text{IR}}=0.50$ and falls back to \res{0} when the rewrite is rejected.}
\label{tab:ir-only-result}
\end{table}

We study how the claim surface (\res{0}--\res{4}) affects evidence retrieval. We index a \texttt{37} GB Wikipedia snapshot (\texttt{2019-08-01}) into \texttt{100}-token passages with a stride of \texttt{50}, matching the dump used to construct DialFact \cite{gupta-etal-2022-dialfact} and ensuring that all annotated gold evidence is, in principle, retrievable from our index.

Since all passages have roughly the same length, we use a relatively weak length normalisation ($b{=}0.4$) and set the term-frequency saturation parameter to $k_1{=}1.5$, following common BM25 settings in Pyserini/BEIR-style benchmarks \cite{lin2021pyserinieasytousepythontoolkit}. Our retrieval pipeline has three steps.

First, BM25 retrieves the top 300 passages per query. Based on an elbow analysis, we fix $K{=}180$ as the working cut-off and pass these 180 candidates to the dense retriever. Second, we apply an E5-large bi-encoder \cite{wang2024multilinguale5textembeddings} to the $K{=}180$ BM25 candidates, retrieve the top 20 semantically similar passages and keep the top 10. Finally, we apply a BGE-reranker-large cross-encoder \cite{chen2024bgem3embeddingmultilingualmultifunctionality} to these 10 passages and select a single passage as the final evidence for IR-only evaluation.

The original claim \res{0} already retrieves most gold evidence (73.89\% R@180 and 48.37\% nDCG@180 with BM25), and the intermediate normalisation variants \res{1}--\res{3} leave IR almost unchanged: all metrics remain within 0.7 points of \res{0} at every stage. Relative to \res{0}, \res{4} improves BM25 recall@180 from 73.89\% to 76.54\% and nDCG@180 from 48.37\% to 50.94\%; E5 dense retrieval recall@10 rises from 70.47\% to 73.18\% and nDCG@10 from 61.98\% to 64.45\%; and BGE cross-encoder recall@1 increases from 53.51\% to 55.53\% with a similar gain in nDCG@1. These gains indicate that making colloquial claims more self-contained by resolving in-scope pronouns helps all three retrieval stages focus on the correct entities.

The gated variant uses the IR threshold $\tau_{\text{IR}}{=}0.50$ chosen in \S\ref{sec:gate-parameters}. Figure~\ref{fig:valid_ir_metrics_t} and Table~\ref{tab:ir-only-result} show that this threshold has almost no effect on retrieval: across BM25, E5, and BGE-CE, \textit{Gated} stays within 0.05 points of \res{4} on all recall and nDCG metrics.

Overall, these IR-only results are consistent with H1: minimal surface normalisation (\res{1}--\res{3}) is retrieval-neutral across all three retrieval stages. In contrast, scoped pronoun rewriting (\res{4}) improves retrieval gains, and applying the IR-tuned gate ($\tau_{\text{IR}}{=}0.50$) preserves these gains.

\subsection{FV-only}
\label{sec:fv-only}
\begin{table}[t]
\centering
\small
\resizebox{\columnwidth}{!}{
\begin{tabular}{lcccccc}
\toprule
Claim & Acc & Macro-F1 & F1(S) & F1(R) & F1(NEI) & $\Delta$F1 \\
\midrule
$\res{0}$
  & 62.67 & 61.09 & 45.08 & 76.77 & 61.42 & -- \\
$\res{1}$
  & 62.66 & 61.08 & 45.00 & 76.78 & 61.46 & -0.01 \\
$\res{2}$
  & 62.71 & 61.16 & 45.33 & 76.69 & 61.47 & +0.07 \\
$\res{3}$
  & 62.80 & 61.22 & 45.25 & 76.75 & 61.67 & +0.13 \\
$\res{4}$
  & 63.15 & 61.85 & 47.59 & 76.20 & 61.77 & +0.76 \\
+\textit{Gated}
  & \textbf{63.77} & \textbf{62.93} & \textbf{50.59} & 76.12 & \textbf{62.10} & \textbf{+1.84} \\
$\res{5}$
  & 58.27 & 56.93 & 42.79 & 67.61 & 60.40 & -4.16 \\
+\textit{Gated}
  & 62.23 & 60.40 & 42.92 & \textbf{77.00} & 61.28 & -0.69 \\
\bottomrule
\end{tabular}
}
\caption{FV-only results on test split with gold evidence, comparing claim surfaces \res{0}-\res{5} (Accuracy and macro/class-wise F1, \%). \res{4}+\textit{Gated} and \res{5}+\textit{Gated} apply BiCon-Gate with $\tau_{\text{FV}}=0.70$ (fallback to \res{0}), and $\Delta$F1 is the macro-F1 change over \res{0}.}
\label{tab:fv-only-result}
\end{table}

We run FV-only with gold evidence and an NLI verifier \cite{he2023debertav3improvingdebertausing, laurer2024buildingefficientuniversalclassifiers}; hypotheses use the last two dialogue turns by default, and \res{4}+\textit{Gated} applies BiCon-Gate with the FV-tuned threshold $\tau_{\text{FV}}{=}0.70$ (see \S\ref{sec:gate-parameters}). Table~\ref{tab:fv-only-result} summarises FV-only performance for \res{0}--\res{5} and the gated variant of \res{4}. The original claim \res{0} attains 61.09\% macro-F1 and 62.67\% accuracy.

Cumulative claim de-colloquialisation (\res{1}--\res{3}) leaves FV almost unchanged when premises are gold evidence: macro-F1 stays within $\pm 0.2$ points of \res{0}, and class-wise F1 for S/R/NEI shifts by at most 0.3 points.

In contrast, scoped coreference rewriting \res{4} yields a noticeable FV gain.
\res{4} improves macro-F1 from 61.09\% to 61.85\% (+0.76) and accuracy from 62.67\% to 63.15\%. Most of this gain comes from the SUPPORTS class: F1(S) rises from 45.08\% to 47.59\%, while F1(NEI) also slightly improves (61.42\% $\rightarrow$ 61.77\%), and F1(R) decreases slightly (76.77\% $\rightarrow$ 76.20\%).

Making the claims more self-contained by replacing in-scope pronouns therefore helps the verifier distinguish supported facts without hurting NEI. This behaviour contrasts with observations by \citet{chamoun-etal-2023-automated}, who report that generic coreference resolution and claim rewriting can improve document recall for conversational claims, but tend to degrade sentence-level evidence selection and claim verification on DialFact due to rewriting and resolution errors. In our setup, the scoped \res{4} rewrites already improve FV over \res{0}.

The gated variant attains the best FV-only performance with 62.93\% macro-F1 (+1.84 over \res{0}) and 63.77\% accuracy, while F1(S) increases further to 50.59\%. For comparison, the decoder one-shot rewrite \res{5} is markedly worse than \res{0} (56.93\% macro-F1), and even with BiCon-Gate it recovers only to 60.40\% macro-F1, motivating the ablation discussion in \S\ref{sec:analysis}. These trends are robust to context window size: Appendix Table~\ref{tab:fv-only-context} and Figures~\ref{fig:heatmap-macro}--\ref{fig:heatmap-classwise} summarise $\Delta$Macro-F1 and class-wise $\Delta$F1 over $k$.

Importantly, we keep the FV threshold fixed at $\tau_{\text{FV}}=0.70$ when varying the context window size $k$. The consistent gains of \res{4}+\textit{Gated} across $k$ in Appendix Table~\ref{tab:fv-only-context} therefore suggest that this operating point is reasonably stable under this moderate distribution shift in dialogue context length.

Overall, these FV-only results are consistent with H1 and H2. With gold evidence provided as premises, light de-colloquialisation (\res{1}--\res{3}) remains largely verification-neutral, while scoped rewriting is the most reliable when filtered by BiCon-Gate.

\subsection{End-to-End}
\label{sec:e2e}
\begin{table}[t]
\centering
\small
\resizebox{\columnwidth}{!}{
\begin{tabular}{lcccccc}
\toprule
Claim & Acc & Macro-F1 & F1(S) & F1(R) & F1(NEI) & $\Delta$F1 \\
\midrule
$\res{0}$
  & 34.85 & 22.60 & 2.45 & 15.99 & 49.37 & 0.00  \\
$\res{1}$
  & 34.86 & 22.62 & 2.50 & 16.01 & 49.37 & +0.02 \\
$\res{2}$
  & 34.96 & 22.71 & 2.55 & 16.13 & 49.45 & +0.11 \\
$\res{3}$
  & 35.02 & 22.88 & 2.89 & \textbf{16.31} & 49.44 & +0.28\\
$\res{4}$
  & 34.10 & 20.73 & 3.03 & 9.57  & 49.60 & $-1.87$  \\
+\textit{Gated}
  &\textbf{ 35.54} & \textbf{23.22} &\textbf{4.56} & 14.84 & \textbf{50.24 }& \textbf{+0.62}  \\
\bottomrule
\end{tabular}
}
\caption{End-to-end results on the full test split using the top-1 retrieved passage as evidence (Accuracy and macro/class-wise F1, \%). \res{4}+\textit{Gated} uses BiCon-Gate with $\tau_{FV}{=}0.70$.}
\label{tab:e2e-result}
\end{table}

For each claim surface \res{k}, we run the IR pipeline from \S\ref{sec:ir-only} and take the BGE cross-encoder's top-1 passage as the premise. We then apply the NLI verifier from \S\ref{sec:fv-only}, constructing the hypothesis exactly as in FV-only (the last two turns in context followed by \res{k}). Since DialFact provides 1.32 gold evidence items per claim on average (test factual subset), top-1 is a conservative but practical choice that isolates claim-surface effects under retrieval noise; however, it may underestimate gains achievable with multi-evidence aggregation (Table~\ref{tab:e2e-result}).

Table~\ref{tab:e2e-result} shows that using \res{0} yields 22.60 macro-F1 and 34.85\% accuracy, substantially lower than in the FV-only setting with gold evidence, reflecting the difficulty of relying on a single retrieved passage. Light de-colloquialisation (\res{1}--\res{3}) has only a small impact on the full pipeline. Macro-F1 remains within 0.3 points of the \res{0} (22.60 $\rightarrow$ 22.88 for \res{3}), and accuracy increases only slightly from 34.85\% to 35.02\% for \res{3}.

Class-wise F1 follows a similar pattern across \res{\text{0-3}}: NEI stays high and stable ($\approx 49.3\text{-}49.5$), while SUPPORTS and REFUTES remain much lower (F1(S) $\leq 2.9$, F1(R) $\approx 16$). This suggests that, once retrieval noise is introduced, the main bottleneck is deciding SUPPORTS vs. REFUTES under noisy top-1 evidence, rather than the exact surface form of the claim.

Scoped coreference rewriting \res{4} shows a different trade-off. Although \res{4} improves document-level IR metrics (\S\ref{sec:ir-only}), it hurts E2E performance: macro-F1 drops from 22.60 to 20.73 and accuracy from 34.85 to 34.10. Most of this loss is due to REFUTES: F1(R) falls from 16.31 for \res{3} to 9.57 for \res{4}, while F1(NEI) stays essentially unchanged (49.60\%).

Applying BiCon-Gate on top of \res{4} largely recovers these losses: the gated variant (\res{4}+\textit{Gated}) uses the FV-tuned threshold from \S\ref{sec:gate-parameters} to decide, for each example, whether to use \res{4} (if $s_i^{(4)} \ge \tau_{\text{FV}}$) or fall back to \res{0} otherwise. We retrieve evidence using the same claim surface selected by the gate. This simple policy yields the best E2E performance: macro-F1 rises to 23.22 (+0.62 over \res{0}) and accuracy to 35.54\% (+0.69). Class-wise, the gate boosts F1(S) to 4.56 and raises F1(R) back to 14.84.
Taken together, these E2E results are consistent with H3: by selectively accepting rewrites and otherwise falling back to \res{0}, BiCon-Gate mitigates the rewrite-induced IR--FV trade-off and yields the best E2E performance.

\subsection{Analysis}
\label{sec:analysis}
This section provides three analyses of BiCon-Gate: (i) gate activation rates at the IR and FV/E2E operating points, clarifying how often the pipeline applies the scoped rewrite (\res{4}) rather than falling back to the original claim (\res{0}); (ii) an ablation comparing the decoder-based one-shot rewrite (\res{5}) to our scoped coreference rewrite (\res{4}) to isolate the impact of aggressive reformulation on verification; and (iii) qualitative error patterns that explain when rewriting helps and when it introduces semantic drift.

\paragraph{Gate activation rates.}
\label{sec:passrate}
We report the routing (activation) rates at the IR and FV/E2E operating points. At $\tau_{\text{IR}}{=}0.50$, the gate routes 52.59\% of factual test queries to \res{4}.

At $\tau_{\text{FV}}{=}0.70$, it routes 46.56\% of test instances to \res{4}. In the \res{5} ablation at the same FV threshold, the gate accepts only 0.92\% of \res{5} candidates (mean score 0.52), so \res{5}+\textit{Gated} almost always falls back to \res{0}. BiCon-Gate acts as a semantic router, sending high-confidence scoped rewrites to \res{4} and routing drift-prone paraphrases back to \res{0}.

\paragraph{Decoder-based rewriting vs. scoped coreference}
\label{sec:r5-ablation}
\begin{figure}[t]
    \centering
    \includegraphics[width=\linewidth,clip]{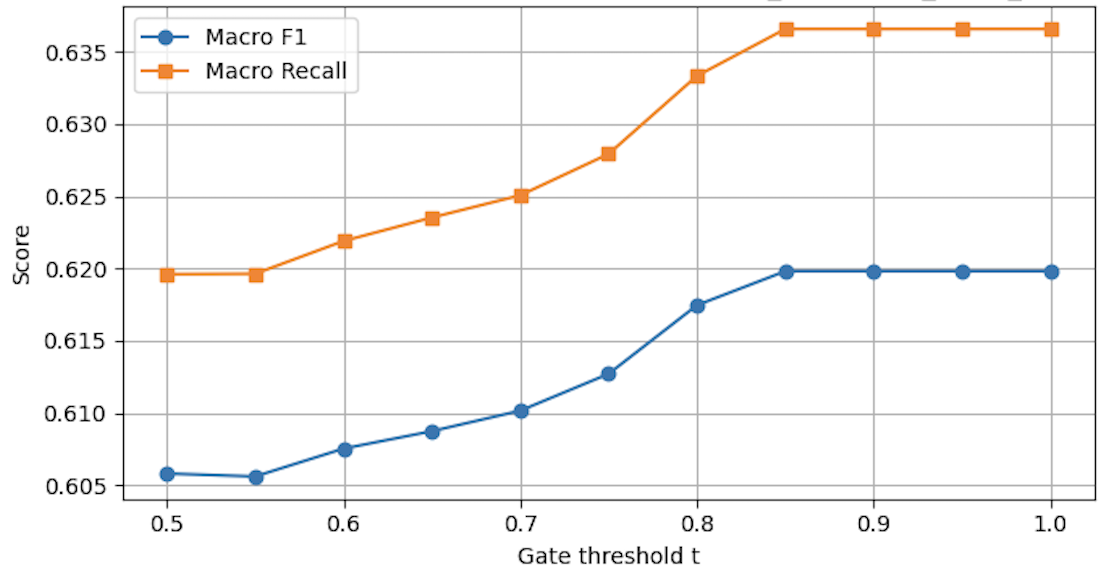}
    \caption{
        Effect of gating on the decoder one-shot rewrite (\res{5}) in FV-only on the validation split. As $\tau$ increases (rejecting more \res{5} instances), macro-F1 and macro-Recall approach the \res{0} baseline, indicating that the gate mitigates noisy rewrites primarily via fallback to \res{0}.
    }
    \label{fig:valid_f1_t_r5}
\end{figure}

We compare the decoder one-shot rewrite \res{5} (Table~\ref{tab:prompt}) against the scoped pronoun rewrite \res{4}, which performs a targeted antecedent substitution only when a pronoun is in-scope (i.e., its antecedent appears in the dialogue context $C$).
Across context lengths ($k\ge2$), \res{4} consistently improves FV performance (best with \res{4}+\textit{Gated}), whereas \res{5} degrades FV (Appendix Table~\ref{tab:fv-only-context}).

Table~\ref{tab:fv-only-result} highlights the contrast clearly: while scoped rewriting \res{4} improves FV (61.85\% macro-F1) and its gated version further gains to 62.93\%, the decoder rewrite \res{5} collapses to 56.93\% and its gated recovers only 60.40\%. Consistent with the low acceptance rate for \res{5} at $\tau_{\text{FV}}{=}0.70$ (\S\ref{sec:passrate}), this gap suggests that one-shot rewrites frequently introduce semantic drift, whereas targeted edits can be useful when selectively accepted.

Class-wise analysis supports this interpretation. Appendix Figure~\ref{fig:heatmap-classwise} shows that \res{5}'s harm is driven primarily by a large drop in REFUTES across $k$, while \res{4}+\textit{Gated} improves mostly via SUPPORTS gains.
Further, Figure~\ref{fig:valid_f1_t_r5} shows that \res{5} appears to improve mainly as the threshold increases and the system increasingly falls back toward \res{0}, whereas \res{4} exhibits a clear interior optimum (Figure~\ref{fig:valid_f1_t_r4}), indicating that a non-trivial subset of high-confidence scoped rewrites is genuinely beneficial.

\paragraph{Qualitative Error Analysis.}
\label{sec:error-analysis}

Appendix Table~\ref{tab:rewrites} provides representative \res{0}/\res{4}/\res{5} triples; we summarise the recurring patterns here. \res{4} is typically a minimal, semantics-preserving edit: it replaces an in-scope pronoun with its referent from the preceding text $C$ (e.g., \textit{it} $\rightarrow$ \textit{Heartbreak Hotel}, \textit{He} $\rightarrow$ \textit{Elvis}), making the hypothesis more self-contained without broader paraphrasing. This aligns with the consistent SUPPORTS gains observed for \res{4}+\textit{Gated}.

In contrast, \res{5} often performs surface-level normalisation (e.g., true-casing, punctuation, quoting) while leaving context-dependent pronouns unresolved; when it does paraphrase, it can blur cues that matter for contradiction, consistent with its REFUTES degradation (Appendix Figure~\ref{fig:heatmap-classwise}).
Gate scores reflect this difference: in these examples, \res{4}+\textit{Gated} receives consistently higher scores than \res{5}+\textit{Gated} (mean 0.79 vs. 0.54; Appendix Table~\ref{tab:rewrites}), suggesting that BiCon-Gate favours verification-oriented de-contextualisation over cosmetic or potentially drifting rewrites.

\section{Conclusions}

We study how colloquial, context-dependent dialogue claims degrade both retrieval and verification, and propose a conservative de-colloquialisation pipeline---lightweight surface normalisation and scoped in-claim pronominal rewriting---paired with BiCon-Gate, a bidirectional NLI-based router that accepts rewrites only when semantically supported and otherwise falls back to the original claim.

On DialFact, surface-level normalisation is largely neutral, while scoped pronoun rewriting improves document retrieval across sparse, dense, and cross-encoder stages; yet in a strict top-1 E2E setting, ungated coreference rewrites can hurt verification despite better retrieval, highlighting a rewrite-induced IR--FV trade-off under retrieval noise.

BiCon-Gate mitigates this trade-off by selectively accepting high-confidence rewrites, yielding the strongest FV-only gains (notably on SUPPORTS) and the best top-1 E2E performance among the claim variants, whereas one-shot decoder rewrites are less reliable and often harm verification.

Overall, our results support H1--H3: minimal normalisation yields stable IR and FV with only marginal changes; scoped pronominal rewriting improves FV and BiCon-Gate further improves robustness by filtering rewrites that are not semantically supported; and semantic gating mitigates the IR--FV trade-off, making conservative rewriting a robust, controllable component in retrieval--verification pipelines.

\clearpage
\section*{Limitations}
First, we evaluate only on the DialFact dataset using an English Wikipedia snapshot. The extent to which the observed gains transfer to other dialogue genres, longer contexts, or languages with different pronominal and morphological systems remains to be validated.

Second, BiCon-Gate depends on multiple off-the-shelf components (a coreference resolver, a retriever, NLI model, and sentence encoders). Their biases, errors, and calibration properties can affect gate decisions; moreover, gate thresholds tuned on DialFact may require retuning under distribution shifts, and the resulting multi-model pipeline increases complexity, latency, and inference cost.

Third, the scope of rewriting is intentionally narrow---limited to light normalisation and in-scope pronominal resolution---leaving other colloquial phenomena (e.g., deixis, ellipsis, or filler words) unaddressed. In addition, incorrect antecedent substitutions can introduce errors that propagate to downstream retrieval and verification.

Finally, our IR is passage-level and E2E setting uses a fixed retriever stack (BM25 $\rightarrow$ E5 $\rightarrow$ BGE-CE) and only the top-1 retrieved passage as evidence. Results may differ with multi-passage evidence aggregation, alternative retriever-verifier architectures, joint training, or recent evidence collections.

\section*{Ethics Statement}
We use a publicly available dataset (DialFact) and an English Wikipedia snapshot, along with publicly released pretrained models, and we do not involve new data collection or human-subject studies. Our rewriting step---especially pronoun resolution---may introduce factual or attribution errors, which can lead to incorrect retrieval and verification outcomes. We therefore report aggregate benchmark results and caution against using outputs as definitive factual judgments without human oversight and transparent access to supporting evidence.

\bibliography{acl}

@inproceedings{pisarevskaya2025zeroshotfewshotlearninginstructionfollowing,
  title={Zero-shot and Few-shot Learning with Instruction-following LLMs for Claim Matching in Automated Fact-checking},
  author={Pisarevskaya, Dina and Zubiaga, Arkaitz},
  booktitle={Proceedings of the 31st International Conference on Computational Linguistics},
  pages={9721--9736},
  year={2025}
}

@inproceedings{thorne-vlachos-2018-automated,
  title = "Automated Fact Checking: Task Formulations, Methods and Future Directions",
  author = "Thorne, James  and
      Vlachos, Andreas",
  editor = "Bender, Emily M.  and
      Derczynski, Leon  and
      Isabelle, Pierre",
  booktitle = "Proceedings of the 27th International Conference on Computational Linguistics",
  month = aug,
  year = "2018",
  address = "Santa Fe, New Mexico, USA",
  publisher = "Association for Computational Linguistics",
  url = "https://aclanthology.org/C18-1283/",
  pages = "3346--3359"
}

@misc{chen2024bgem3embeddingmultilingualmultifunctionality,
  title={BGE M3-Embedding: Multi-Lingual, Multi-Functionality, Multi-Granularity Text Embeddings Through Self-Knowledge Distillation},
  author={Jianlv Chen and Shitao Xiao and Peitian Zhang and Kun Luo and Defu Lian and Zheng Liu},
  year={2024},
  eprint={2402.03216},
  archivePrefix={arXiv},
  primaryClass={cs.CL},
  url={https://arxiv.org/abs/2402.03216}
}

@misc{wang2024multilinguale5textembeddings,
  title={Multilingual E5 Text Embeddings: A Technical Report},
  author={Liang Wang and Nan Yang and Xiaolong Huang and Linjun Yang and Rangan Majumder and Furu Wei},
  year={2024},
  eprint={2402.05672},
  archivePrefix={arXiv},
  primaryClass={cs.CL},
  url={https://arxiv.org/abs/2402.05672}
}

@misc{laurer2024buildingefficientuniversalclassifiers,
  title={Building Efficient Universal Classifiers with Natural Language Inference},
  author={Moritz Laurer and Wouter van Atteveldt and Andreu Casas and Kasper Welbers},
  year={2024},
  eprint={2312.17543},
  archivePrefix={arXiv},
  primaryClass={cs.CL},
  url={https://arxiv.org/abs/2312.17543}
}

@misc{he2023debertav3improvingdebertausing,
  title={DeBERTaV3: Improving DeBERTa using ELECTRA-Style Pre-Training with Gradient-Disentangled Embedding Sharing},
  author={Pengcheng He and Jianfeng Gao and Weizhu Chen},
  year={2023},
  eprint={2111.09543},
  archivePrefix={arXiv},
  primaryClass={cs.CL},
  url={https://arxiv.org/abs/2111.09543}
}

@misc{grattafiori2024llama3herdmodels,
  title={The Llama 3 Herd of Models},
  author={Aaron Grattafiori and Abhimanyu Dubey and Abhinav Jauhri and Abhinav Pandey and Abhishek Kadian and Ahmad Al-Dahle and Aiesha Letman and Akhil Mathur and Alan Schelten and Alex Vaughan and Amy Yang and Angela Fan and Anirudh Goyal and Anthony Hartshorn and Aobo Yang and Archi Mitra and Archie Sravankumar and Artem Korenev and Arthur Hinsvark and Arun Rao and Aston Zhang and Aurelien Rodriguez and Austen Gregerson and Ava Spataru and Baptiste Roziere and Bethany Biron and Binh Tang and Bobbie Chern and Charlotte Caucheteux and Chaya Nayak and Chloe Bi and Chris Marra and Chris McConnell and Christian Keller and Christophe Touret and Chunyang Wu and Corinne Wong and Cristian Canton Ferrer and Cyrus Nikolaidis and Damien Allonsius and Daniel Song and Danielle Pintz and Danny Livshits and Danny Wyatt and David Esiobu and Dhruv Choudhary and Dhruv Mahajan and Diego Garcia-Olano and Diego Perino and Dieuwke Hupkes and Egor Lakomkin and Ehab AlBadawy and Elina Lobanova and Emily Dinan and Eric Michael Smith and Filip Radenovic and Francisco Guzmán and Frank Zhang and Gabriel Synnaeve and Gabrielle Lee and Georgia Lewis Anderson and Govind Thattai and Graeme Nail and Gregoire Mialon and Guan Pang and Guillem Cucurell and Hailey Nguyen and Hannah Korevaar and Hu Xu and Hugo Touvron and Iliyan Zarov and Imanol Arrieta Ibarra and Isabel Kloumann and Ishan Misra and Ivan Evtimov and Jack Zhang and Jade Copet and Jaewon Lee and Jan Geffert and Jana Vranes and Jason Park and Jay Mahadeokar and Jeet Shah and Jelmer van der Linde and Jennifer Billock and Jenny Hong and Jenya Lee and Jeremy Fu and Jianfeng Chi and Jianyu Huang and Jiawen Liu and Jie Wang and Jiecao Yu and Joanna Bitton and Joe Spisak and Jongsoo Park and Joseph Rocca and Joshua Johnstun and Joshua Saxe and Junteng Jia and Kalyan Vasuden Alwala and Karthik Prasad and Kartikeya Upasani and Kate Plawiak and Ke Li and Kenneth Heafield and Kevin Stone and Khalid El-Arini and Krithika Iyer and Kshitiz Malik and Kuenley Chiu and Kunal Bhalla and Kushal Lakhotia and Lauren Rantala-Yeary and Laurens van der Maaten and Lawrence Chen and Liang Tan and Liz Jenkins and Louis Martin and Lovish Madaan and Lubo Malo and Lukas Blecher and Lukas Landzaat and Luke de Oliveira and Madeline Muzzi and Mahesh Pasupuleti and Mannat Singh and Manohar Paluri and Marcin Kardas and Maria Tsimpoukelli and Mathew Oldham and Mathieu Rita and Maya Pavlova and Melanie Kambadur and Mike Lewis and Min Si and Mitesh Kumar Singh and Mona Hassan and Naman Goyal and Narjes Torabi and Nikolay Bashlykov and Nikolay Bogoychev and Niladri Chatterji and Ning Zhang and Olivier Duchenne and Onur Çelebi and Patrick Alrassy and Pengchuan Zhang and Pengwei Li and Petar Vasic and Peter Weng and Prajjwal Bhargava and Pratik Dubal and Praveen Krishnan and Punit Singh Koura and Puxin Xu and Qing He and Qingxiao Dong and Ragavan Srinivasan and Raj Ganapathy and Ramon Calderer and Ricardo Silveira Cabral and Robert Stojnic and Roberta Raileanu and Rohan Maheswari and Rohit Girdhar and Rohit Patel and Romain Sauvestre and Ronnie Polidoro and Roshan Sumbaly and Ross Taylor and Ruan Silva and Rui Hou and Rui Wang and Saghar Hosseini and Sahana Chennabasappa and Sanjay Singh and Sean Bell and Seohyun Sonia Kim and Sergey Edunov and Shaoliang Nie and Sharan Narang and Sharath Raparthy and Sheng Shen and Shengye Wan and Shruti Bhosale and Shun Zhang and Simon Vandenhende and Soumya Batra and Spencer Whitman and Sten Sootla and Stephane Collot and Suchin Gururangan and Sydney Borodinsky and Tamar Herman and Tara Fowler and Tarek Sheasha and Thomas Georgiou and Thomas Scialom and Tobias Speckbacher and Todor Mihaylov and Tong Xiao and Ujjwal Karn and Vedanuj Goswami and Vibhor Gupta and Vignesh Ramanathan and Viktor Kerkez and Vincent Gonguet and Virginie Do and Vish Vogeti and Vítor Albiero and Vladan Petrovic and Weiwei Chu and Wenhan Xiong and Wenyin Fu and Whitney Meers and Xavier Martinet and Xiaodong Wang and Xiaofang Wang and Xiaoqing Ellen Tan and Xide Xia and Xinfeng Xie and Xuchao Jia and Xuewei Wang and Yaelle Goldschlag and Yashesh Gaur and Yasmine Babaei and Yi Wen and Yiwen Song and Yuchen Zhang and Yue Li and Yuning Mao and Zacharie Delpierre Coudert and Zheng Yan and Zhengxing Chen and Zoe Papakipos and Aaditya Singh and Aayushi Srivastava and Abha Jain and Adam Kelsey and Adam Shajnfeld and Adithya Gangidi and Adolfo Victoria and Ahuva Goldstand and Ajay Menon and Ajay Sharma and Alex Boesenberg and Alexei Baevski and Allie Feinstein and Amanda Kallet and Amit Sangani and Amos Teo and Anam Yunus and Andrei Lupu and Andres Alvarado and Andrew Caples and Andrew Gu and Andrew Ho and Andrew Poulton and Andrew Ryan and Ankit Ramchandani and Annie Dong and Annie Franco and Anuj Goyal and Aparajita Saraf and Arkabandhu Chowdhury and Ashley Gabriel and Ashwin Bharambe and Assaf Eisenman and Azadeh Yazdan and Beau James and Ben Maurer and Benjamin Leonhardi and Bernie Huang and Beth Loyd and Beto De Paola and Bhargavi Paranjape and Bing Liu and Bo Wu and Boyu Ni and Braden Hancock and Bram Wasti and Brandon Spence and Brani Stojkovic and Brian Gamido and Britt Montalvo and Carl Parker and Carly Burton and Catalina Mejia and Ce Liu and Changhan Wang and Changkyu Kim and Chao Zhou and Chester Hu and Ching-Hsiang Chu and Chris Cai and Chris Tindal and Christoph Feichtenhofer and Cynthia Gao and Damon Civin and Dana Beaty and Daniel Kreymer and Daniel Li and David Adkins and David Xu and Davide Testuggine and Delia David and Devi Parikh and Diana Liskovich and Didem Foss and Dingkang Wang and Duc Le and Dustin Holland and Edward Dowling and Eissa Jamil and Elaine Montgomery and Eleonora Presani and Emily Hahn and Emily Wood and Eric-Tuan Le and Erik Brinkman and Esteban Arcaute and Evan Dunbar and Evan Smothers and Fei Sun and Felix Kreuk and Feng Tian and Filippos Kokkinos and Firat Ozgenel and Francesco Caggioni and Frank Kanayet and Frank Seide and Gabriela Medina Florez and Gabriella Schwarz and Gada Badeer and Georgia Swee and Gil Halpern and Grant Herman and Grigory Sizov and Guangyi and Zhang and Guna Lakshminarayanan and Hakan Inan and Hamid Shojanazeri and Han Zou and Hannah Wang and Hanwen Zha and Haroun Habeeb and Harrison Rudolph and Helen Suk and Henry Aspegren and Hunter Goldman and Hongyuan Zhan and Ibrahim Damlaj and Igor Molybog and Igor Tufanov and Ilias Leontiadis and Irina-Elena Veliche and Itai Gat and Jake Weissman and James Geboski and James Kohli and Janice Lam and Japhet Asher and Jean-Baptiste Gaya and Jeff Marcus and Jeff Tang and Jennifer Chan and Jenny Zhen and Jeremy Reizenstein and Jeremy Teboul and Jessica Zhong and Jian Jin and Jingyi Yang and Joe Cummings and Jon Carvill and Jon Shepard and Jonathan McPhie and Jonathan Torres and Josh Ginsburg and Junjie Wang and Kai Wu and Kam Hou U and Karan Saxena and Kartikay Khandelwal and Katayoun Zand and Kathy Matosich and Kaushik Veeraraghavan and Kelly Michelena and Keqian Li and Kiran Jagadeesh and Kun Huang and Kunal Chawla and Kyle Huang and Lailin Chen and Lakshya Garg and Lavender A and Leandro Silva and Lee Bell and Lei Zhang and Liangpeng Guo and Licheng Yu and Liron Moshkovich and Luca Wehrstedt and Madian Khabsa and Manav Avalani and Manish Bhatt and Martynas Mankus and Matan Hasson and Matthew Lennie and Matthias Reso and Maxim Groshev and Maxim Naumov and Maya Lathi and Meghan Keneally and Miao Liu and Michael L. Seltzer and Michal Valko and Michelle Restrepo and Mihir Patel and Mik Vyatskov and Mikayel Samvelyan and Mike Clark and Mike Macey and Mike Wang and Miquel Jubert Hermoso and Mo Metanat and Mohammad Rastegari and Munish Bansal and Nandhini Santhanam and Natascha Parks and Natasha White and Navyata Bawa and Nayan Singhal and Nick Egebo and Nicolas Usunier and Nikhil Mehta and Nikolay Pavlovich Laptev and Ning Dong and Norman Cheng and Oleg Chernoguz and Olivia Hart and Omkar Salpekar and Ozlem Kalinli and Parkin Kent and Parth Parekh and Paul Saab and Pavan Balaji and Pedro Rittner and Philip Bontrager and Pierre Roux and Piotr Dollar and Polina Zvyagina and Prashant Ratanchandani and Pritish Yuvraj and Qian Liang and Rachad Alao and Rachel Rodriguez and Rafi Ayub and Raghotham Murthy and Raghu Nayani and Rahul Mitra and Rangaprabhu Parthasarathy and Raymond Li and Rebekkah Hogan and Robin Battey and Rocky Wang and Russ Howes and Ruty Rinott and Sachin Mehta and Sachin Siby and Sai Jayesh Bondu and Samyak Datta and Sara Chugh and Sara Hunt and Sargun Dhillon and Sasha Sidorov and Satadru Pan and Saurabh Mahajan and Saurabh Verma and Seiji Yamamoto and Sharadh Ramaswamy and Shaun Lindsay and Shaun Lindsay and Sheng Feng and Shenghao Lin and Shengxin Cindy Zha and Shishir Patil and Shiva Shankar and Shuqiang Zhang and Shuqiang Zhang and Sinong Wang and Sneha Agarwal and Soji Sajuyigbe and Soumith Chintala and Stephanie Max and Stephen Chen and Steve Kehoe and Steve Satterfield and Sudarshan Govindaprasad and Sumit Gupta and Summer Deng and Sungmin Cho and Sunny Virk and Suraj Subramanian and Sy Choudhury and Sydney Goldman and Tal Remez and Tamar Glaser and Tamara Best and Thilo Koehler and Thomas Robinson and Tianhe Li and Tianjun Zhang and Tim Matthews and Timothy Chou and Tzook Shaked and Varun Vontimitta and Victoria Ajayi and Victoria Montanez and Vijai Mohan and Vinay Satish Kumar and Vishal Mangla and Vlad Ionescu and Vlad Poenaru and Vlad Tiberiu Mihailescu and Vladimir Ivanov and Wei Li and Wenchen Wang and Wenwen Jiang and Wes Bouaziz and Will Constable and Xiaocheng Tang and Xiaojian Wu and Xiaolan Wang and Xilun Wu and Xinbo Gao and Yaniv Kleinman and Yanjun Chen and Ye Hu and Ye Jia and Ye Qi and Yenda Li and Yilin Zhang and Ying Zhang and Yossi Adi and Youngjin Nam and Yu and Wang and Yu Zhao and Yuchen Hao and Yundi Qian and Yunlu Li and Yuzi He and Zach Rait and Zachary DeVito and Zef Rosnbrick and Zhaoduo Wen and Zhenyu Yang and Zhiwei Zhao and Zhiyu Ma},
  year={2024},
  eprint={2407.21783},
  archivePrefix={arXiv},
  primaryClass={cs.AI},
  url={https://arxiv.org/abs/2407.21783}
}

@inproceedings{devlin2019bertpretrainingdeepbidirectional,
  title={{BERT}: Pre-training of deep bidirectional transformers for language understanding},
  author={Devlin, Jacob and Chang, Ming-Wei and Lee, Kenton and Toutanova, Kristina},
  booktitle={Proceedings of the 2019 conference of the North American chapter of the association for computational linguistics: human language technologies, volume 1 (long and short papers)},
  pages={4171--4186},
  year={2019}
}

@article{vandeghinste2023fullstoppunctuationsegmentationpredictiondutch,
  title={Fullstop: Punctuation and segmentation prediction for dutch with transformers},
  author={Vandeghinste, Vincent and Guhr, Oliver},
  journal={Language Resources and Evaluation},
  volume={58},
  number={4},
  pages={1335--1354},
  year={2024},
  publisher={Springer}
}

@inproceedings{thorne-etal-2018-fever,
  title = "{FEVER}: a Large-scale Dataset for Fact Extraction and {VER}ification",
  author = "Thorne, James  and
      Vlachos, Andreas  and
      Christodoulopoulos, Christos  and
      Mittal, Arpit",
  editor = "Walker, Marilyn  and
      Ji, Heng  and
      Stent, Amanda",
  booktitle = "Proceedings of the 2018 Conference of the North {A}merican Chapter of the Association for Computational Linguistics: Human Language Technologies, Volume 1 (Long Papers)",
  month = jun,
  year = "2018",
  address = "New Orleans, Louisiana",
  publisher = "Association for Computational Linguistics",
  url = "https://aclanthology.org/N18-1074/",
  doi = "10.18653/v1/N18-1074",
  pages = "809--819"
}

@inproceedings{gan-etal-2024-assessing,
  title = "Assessing the Capabilities of Large Language Models in Coreference: An Evaluation",
  author = "Gan, Yujian  and
      Poesio, Massimo  and
      Yu, Juntao",
  editor = "Calzolari, Nicoletta  and
      Kan, Min-Yen  and
      Hoste, Veronique  and
      Lenci, Alessandro  and
      Sakti, Sakriani  and
      Xue, Nianwen",
  booktitle = "Proceedings of the 2024 Joint International Conference on Computational Linguistics, Language Resources and Evaluation (LREC-COLING 2024)",
  month = may,
  year = "2024",
  address = "Torino, Italia",
  publisher = "ELRA and ICCL",
  url = "https://aclanthology.org/2024.lrec-main.145/",
  pages = "1645--1665"
}

@inproceedings{cao-etal-2024-incomplete,
  title = "Incomplete Utterance Rewriting with Editing Operation Guidance and Utterance Augmentation",
  author = "Cao, Zhiyu  and
      Li, Peifeng  and
      Fan, Yaxin  and
      Zhu, Qiaoming",
  editor = "Al-Onaizan, Yaser  and
      Bansal, Mohit  and
      Chen, Yun-Nung",
  booktitle = "Proceedings of the 2024 Conference on Empirical Methods in Natural Language Processing",
  month = nov,
  year = "2024",
  address = "Miami, Florida, USA",
  publisher = "Association for Computational Linguistics",
  url = "https://aclanthology.org/2024.emnlp-main.410/",
  doi = "10.18653/v1/2024.emnlp-main.410",
  pages = "7225--7238"
}

@misc{hui2024qwen2,
  title={Qwen2.5 Technical Report},
  author={Qwen and An Yang and Baosong Yang and Beichen Zhang and Binyuan Hui and Bo Zheng and Bowen Yu and Chengyuan Li and Dayiheng Liu and Fei Huang and Haoran Wei and Huan Lin and Jian Yang and Jianhong Tu and Jianwei Zhang and Jianxin Yang and Jiaxi Yang and Jingren Zhou and Junyang Lin and Kai Dang and Keming Lu and Keqin Bao and Kexin Yang and Le Yu and Mei Li and Mingfeng Xue and Pei Zhang and Qin Zhu and Rui Men and Runji Lin and Tianhao Li and Tianyi Tang and Tingyu Xia and Xingzhang Ren and Xuancheng Ren and Yang Fan and Yang Su and Yichang Zhang and Yu Wan and Yuqiong Liu and Zeyu Cui and Zhenru Zhang and Zihan Qiu},
  year={2025},
  eprint={2412.15115},
  archivePrefix={arXiv},
  primaryClass={cs.CL},
  url={https://arxiv.org/abs/2412.15115}
}

@inproceedings{guo-etal-2024-context,
  title = "Context-Aware Tracking and Dynamic Introduction for Incomplete Utterance Rewriting in Extended Multi-Turn Dialogues",
  author = "Guo, Xinnan  and
      Zhu, Qian  and
      Shi, Qiuhui  and
      Lin, Xuan  and
      Wang, Liubin  and
      DaqianLi, DaqianLi  and
      Chen, Yongrui",
  editor = "Ku, Lun-Wei  and
      Martins, Andre  and
      Srikumar, Vivek",
  booktitle = "Findings of the Association for Computational Linguistics: ACL 2024",
  month = aug,
  year = "2024",
  address = "Bangkok, Thailand",
  publisher = "Association for Computational Linguistics",
  url = "https://aclanthology.org/2024.findings-acl.127/",
  doi = "10.18653/v1/2024.findings-acl.127",
  pages = "2138--2148"
}

@inproceedings{li-etal-2023-incomplete,
  title = "Incomplete Utterance Rewriting by A Two-Phase Locate-and-Fill Regime",
  author = "Li, Zitong  and
      Li, Jiawei  and
      Tang, Haifeng  and
      Zhu, Kenny  and
      Yang, Ruolan",
  editor = "Rogers, Anna  and
      Boyd-Graber, Jordan  and
      Okazaki, Naoaki",
  booktitle = "Findings of the Association for Computational Linguistics: ACL 2023",
  month = jul,
  year = "2023",
  address = "Toronto, Canada",
  publisher = "Association for Computational Linguistics",
  url = "https://aclanthology.org/2023.findings-acl.171/",
  doi = "10.18653/v1/2023.findings-acl.171",
  pages = "2731--2745"
}

@inproceedings{deng-etal-2024-document,
  title = "Document-level Claim Extraction and Decontextualisation for Fact-Checking",
  author = "Deng, Zhenyun  and
      Schlichtkrull, Michael  and
      Vlachos, Andreas",
  editor = "Ku, Lun-Wei  and
      Martins, Andre  and
      Srikumar, Vivek",
  booktitle = "Proceedings of the 62nd Annual Meeting of the Association for Computational Linguistics (Volume 1: Long Papers)",
  month = aug,
  year = "2024",
  address = "Bangkok, Thailand",
  publisher = "Association for Computational Linguistics",
  url = "https://aclanthology.org/2024.acl-long.645/",
  doi = "10.18653/v1/2024.acl-long.645",
  pages = "11943--11954"
}

@inproceedings{sundriyal-etal-2023-chaos,
  title = "From Chaos to Clarity: Claim Normalization to Empower Fact-Checking",
  author = "Sundriyal, Megha  and
      Chakraborty, Tanmoy  and
      Nakov, Preslav",
  editor = "Bouamor, Houda  and
      Pino, Juan  and
      Bali, Kalika",
  booktitle = "Findings of the Association for Computational Linguistics: EMNLP 2023",
  month = dec,
  year = "2023",
  address = "Singapore",
  publisher = "Association for Computational Linguistics",
  url = "https://aclanthology.org/2023.findings-emnlp.439/",
  doi = "10.18653/v1/2023.findings-emnlp.439",
  pages = "6594--6609"
}

@inproceedings{kim-etal-2021-robust,
  title = "How Robust are Fact Checking Systems on Colloquial Claims?",
  author = "Kim, Byeongchang  and
      Kim, Hyunwoo  and
      Hong, Seokhee  and
      Kim, Gunhee",
  editor = "Toutanova, Kristina  and
      Rumshisky, Anna  and
      Zettlemoyer, Luke  and
      Hakkani-Tur, Dilek  and
      Beltagy, Iz  and
      Bethard, Steven  and
      Cotterell, Ryan  and
      Chakraborty, Tanmoy  and
      Zhou, Yichao",
  booktitle = "Proceedings of the 2021 Conference of the North American Chapter of the Association for Computational Linguistics: Human Language Technologies",
  month = jun,
  year = "2021",
  address = "Online",
  publisher = "Association for Computational Linguistics",
  url = "https://aclanthology.org/2021.naacl-main.121/",
  doi = "10.18653/v1/2021.naacl-main.121",
  pages = "1535--1548"
}

@inproceedings{chamoun-etal-2023-automated,
  title = "Automated Fact-Checking in Dialogue: Are Specialized Models Needed?",
  author = "Chamoun, Eric  and
      Saeidi, Marzieh  and
      Vlachos, Andreas",
  editor = "Bouamor, Houda  and
      Pino, Juan  and
      Bali, Kalika",
  booktitle = "Proceedings of the 2023 Conference on Empirical Methods in Natural Language Processing",
  month = dec,
  year = "2023",
  address = "Singapore",
  publisher = "Association for Computational Linguistics",
  url = "https://aclanthology.org/2023.emnlp-main.993/",
  doi = "10.18653/v1/2023.emnlp-main.993",
  pages = "16009--16020"
}

@inproceedings{martinelli-etal-2024-maverick,
  title = "Maverick: Efficient and Accurate Coreference Resolution Defying Recent Trends",
  author = "Martinelli, Giuliano  and
      Barba, Edoardo  and
      Navigli, Roberto",
  editor = "Ku, Lun-Wei  and
      Martins, Andre  and
      Srikumar, Vivek",
  booktitle = "Proceedings of the 62nd Annual Meeting of the Association for Computational Linguistics (Volume 1: Long Papers)",
  month = aug,
  year = "2024",
  address = "Bangkok, Thailand",
  publisher = "Association for Computational Linguistics",
  url = "https://aclanthology.org/2024.acl-long.722/",
  doi = "10.18653/v1/2024.acl-long.722",
  pages = "13380--13394"
}

@inproceedings{gupta-etal-2022-dialfact,
  title = "{D}ial{F}act: A Benchmark for Fact-Checking in Dialogue",
  author = "Gupta, Prakhar  and
      Wu, Chien-Sheng  and
      Liu, Wenhao  and
      Xiong, Caiming",
  editor = "Muresan, Smaranda  and
      Nakov, Preslav  and
      Villavicencio, Aline",
  booktitle = "Proceedings of the 60th Annual Meeting of the Association for Computational Linguistics (Volume 1: Long Papers)",
  month = may,
  year = "2022",
  address = "Dublin, Ireland",
  publisher = "Association for Computational Linguistics",
  url = "https://aclanthology.org/2022.acl-long.263/",
  doi = "10.18653/v1/2022.acl-long.263",
  pages = "3785--3801"
}

@inproceedings{manikantan-etal-2025-identifyme,
  title = "{I}dentify{M}e: A Challenging Long-Context Mention Resolution Benchmark for {LLM}s",
  author = "Manikantan, Kawshik  and
      Tapaswi, Makarand  and
      Gandhi, Vineet  and
      Toshniwal, Shubham",
  editor = "Chiruzzo, Luis  and
      Ritter, Alan  and
      Wang, Lu",
  booktitle = "Proceedings of the 2025 Conference of the Nations of the Americas Chapter of the Association for Computational Linguistics: Human Language Technologies (Volume 2: Short Papers)",
  month = apr,
  year = "2025",
  address = "Albuquerque, New Mexico",
  publisher = "Association for Computational Linguistics",
  url = "https://aclanthology.org/2025.naacl-short.64/",
  doi = "10.18653/v1/2025.naacl-short.64",
  pages = "768--777",
  ISBN = "979-8-89176-190-2"
}

@inproceedings{zhang-etal-2023-relevance,
  title = "From Relevance to Utility: Evidence Retrieval with Feedback for Fact Verification",
  author = "Zhang, Hengran  and
      Zhang, Ruqing  and
      Guo, Jiafeng  and
      de Rijke, Maarten  and
      Fan, Yixing  and
      Cheng, Xueqi",
  editor = "Bouamor, Houda  and
      Pino, Juan  and
      Bali, Kalika",
  booktitle = "Findings of the Association for Computational Linguistics: EMNLP 2023",
  month = dec,
  year = "2023",
  address = "Singapore",
  publisher = "Association for Computational Linguistics",
  url = "https://aclanthology.org/2023.findings-emnlp.422/",
  doi = "10.18653/v1/2023.findings-emnlp.422",
  pages = "6373--6384"
}

@inproceedings{xie2024calibratinglanguagemodelsadaptive,
  title={Calibrating Language Models with Adaptive Temperature Scaling},
  author={Xie, Johnathan and Chen, Annie and Lee, Yoonho and Mitchell, Eric and Finn, Chelsea},
  booktitle={Proceedings of the 2024 Conference on Empirical Methods in Natural Language Processing},
  pages={18128--18138},
  year={2024}
}

@inproceedings{guo2017calibrationmodernneuralnetworks,
  title={On calibration of modern neural networks},
  author={Guo, Chuan and Pleiss, Geoff and Sun, Yu and Weinberger, Kilian Q},
  booktitle={Proceedings of the 34th International Conference on Machine Learning-Volume 70},
  pages={1321--1330},
  year={2017}
}

@misc{lin2021pyserinieasytousepythontoolkit,
  title={Pyserini: An Easy-to-Use Python Toolkit to Support Replicable IR Research with Sparse and Dense Representations},
  author={Jimmy Lin and Xueguang Ma and Sheng-Chieh Lin and Jheng-Hong Yang and Ronak Pradeep and Rodrigo Nogueira},
  year={2021},
  eprint={2102.10073},
  archivePrefix={arXiv},
  primaryClass={cs.IR},
  url={https://arxiv.org/abs/2102.10073}
}

@article{chen2023menlirobustevaluationmetrics,
  title={Menli: Robust evaluation metrics from natural language inference},
  author={Chen, Yanran and Eger, Steffen},
  journal={Transactions of the Association for Computational Linguistics},
  volume={11},
  pages={804--825},
  year={2023},
  publisher={MIT Press One Broadway, 12th Floor, Cambridge, Massachusetts 02142, USA~…}
}

@inproceedings{zha2023alignscoreevaluatingfactualconsistency,
  title={AlignScore: Evaluating Factual Consistency with A Unified Alignment Function},
  author={Zha, Yuheng and Yang, Yichi and Li, Ruichen and Hu, Zhiting},
  booktitle={Proceedings of the 61st Annual Meeting of the Association for Computational Linguistics (Volume 1: Long Papers)},
  pages={11328--11348},
  year={2023}
}

@article{laban2021summacrevisitingnlibasedmodels,
  title={SummaC: Re-visiting NLI-based models for inconsistency detection in summarization},
  author={Laban, Philippe and Schnabel, Tobias and Bennett, Paul N and Hearst, Marti A},
  journal={Transactions of the Association for Computational Linguistics},
  volume={10},
  pages={163--177},
  year={2022},
  publisher={MIT Press One Rogers Street, Cambridge, MA 02142-1209, USA journals-info~…}
}

\clearpage
\appendix
\setcounter{table}{0}
\setcounter{figure}{0}
\renewcommand{\thetable}{A\arabic{table}}
\renewcommand{\thefigure}{A\arabic{figure}}

\section{Appendix}
\label{sec:appendix}

\subsection{DialFact Dataset Statistics}
\label{sec:dialfact-stats}
\begin{table}[H]
    \centering
    \small
    \resizebox{\columnwidth}{!}{
    \begin{tabular}{llrrrrrrr}
        \toprule
        Split & Type & \# & S & R & NEI & Ev.Item & Turns($C$) & hasPronouns\\
        \midrule
        Valid & factual  &  8,691 & 3,342 & 3,363 & 1,986 & 1.13 & 4.58 & \textbf{3,395} \\
              & personal &  1,745 &     0 &     0 & 1,745 & 1.01 & 4.36 & 747 \\
              & total    & 10,436 & 3,342 & 3,363 & 3,731 & 1.11 & 4.54 & 4,142 \\
        \midrule
        Test  & factual  & 10,420 & 3,939 & 3,935 & 2,546 & 1.32 & 4.35 & \textbf{4,549} \\
              & personal &  1,389 &     0 &     0 & 1,389 & 1.10 & 3.80 & 679 \\
              & total    & 11,809 & 3,939 & 3,935 & 3,935 & 1.29 & 4.28 & 5,228 \\
        \bottomrule
    \end{tabular}
    }
    \caption{DialFact validation/test statistics and label counts. IR metrics are computed on the \textit{factual} subset; FV/E2E use the full label distribution unless stated otherwise. \texttt{hasPronouns} counts samples containing \emph{at least one in-claim pronoun}.}
\label{tab:dialfact-stat}
\end{table}

\subsection{Evaluation Metrics}
\label{sec:metrics}
\begin{table}[H]
\centering
\resizebox{\columnwidth}{!}{
        \small
            \begin{tabular}{ll}
            \toprule
              Protocol & Metrics \\
              \midrule
              IR (gate tuning)
                & micro-Recall@K,\ $1-\mathrm{ZHR}@K$ (BM25@180) \\
              IR (doc-level)
                & macro-Recall@K,\ nDCG@K \\
                & (BM25@180,\ E5@10,\ BGE-CE@1) \\
              FV-only
                & Accuracy,\ macro-F1,\ classwise F1 (S/R/NEI) \\
              End-to-End
                & Same FV metrics as FV-only, using IR-produced evidence \\
              \bottomrule
              \end{tabular}
    }
    \caption{Evaluation metrics: IR (gate tuning) metrics are computed on the valid split to select $\tau_{\mathrm{IR}}$, whereas IR (doc-level), FV-only, and E2E metrics are reported on the test split. $\mathrm{ZHR}$ is the zero-hit rate; the proportion of queries for which no gold passage is retrieved in the top-K list.}
    \label{tab:metrics}
\end{table}

\subsection{Pronoun Scope}
\label{sec:pronoun-scope}
We rewrite only \emph{in-scope} anaphoric pronouns whose antecedents appear in dialogue context $C$. We exclude deictic uses without textual antecedent and expletive \textit{it}.

\begin{compactitem}
    \item \textbf{Anaphora (coreferential, \emph{in-scope})} A pronoun refers back to a preceding entity in $C$. \newline
    \emph{Ex.} ``I have heard that Louis C.K. performed there in the past.'' / ``I did not know that. \underline{He} was pretty funny.'' (\emph{He} $\rightarrow$ Louis C.K.)
    \item \textbf{Deictic (\emph{out-of-scope})} Reference relies on extra-linguistic context, not on $C$. \newline
    \emph{Ex.} ``\underline{This} is amazing.'' (no textual antecedent)
    \item \textbf{Expletive (\emph{out-of-scope})} Non-referential \textit{it} in weather/raising/extraposition. \newline
    \emph{Ex.} ``\underline{It} is raining.'' / ``\underline{It} seems that she left.'' / ``\underline{It} was John who called.'' / ``\underline{It} is important to exercise.''
\end{compactitem}
\newpage
\subsection{Examples of Rewrites}
\label{sec:example-rewrites}

\begin{table}[H]
\begingroup
\raggedright

\setlength{\abovecaptionskip}{10pt}
\setlength{\belowcaptionskip}{0pt}

{\scriptsize
\setlength{\tabcolsep}{2.6pt}
\renewcommand{\arraystretch}{0.92}

\setlength{\aboverulesep}{0.15ex}
\setlength{\belowrulesep}{0.15ex}
\setlength{\cmidrulesep}{0.15ex}
\setlength{\cmidrulekern}{0.15ex}

\begin{tabularx}{\columnwidth}{@{}p{0.1\columnwidth}>{\raggedright\arraybackslash}X@{}}
\toprule

\multicolumn{2}{@{}p{\dimexpr\columnwidth\relax}@{}}{
\texttt{Example 1 (ID: \detokenize{877___8--1})}
}\\
\\

$C$ &
Elvis is great, he was born in 1935. \ldots\ yeah he really brought rock and roll to the masses, thanks elvis. THANKS A LOT DUDE.\\
\addlinespace[0.15em]

\textbf{\res{0}} &
Remember when heartbreak hotel came out? The public hated \chg{it} at first!\\
\addlinespace[0.08em]

\textbf{\res{4}} &
Remember when heartbreak hotel came out? The public hated \chg{heartbreak hotel} at first! (gate$(\res{4}) = 0.8178$)\\
\addlinespace[0.08em]

\textbf{\res{5}} &
Remember when \chg{``Heartbreak Hotel''} came out? The public hated \chg{it} at first! (gate$(\res{5}) = 0.5445$)\\
\addlinespace[0.22em]

\midrule
\multicolumn{2}{@{}p{\dimexpr\columnwidth\relax}@{}}{
\texttt{Example 2 (ID: \detokenize{445___6--2})}
}\\
\\

$C$ &
I love The Walking Dead, I've seen every episode since it premiered on October 31, 2010. \ldots\ Do you know what network I can find The Walking Dead on?\\
\addlinespace[0.15em]

\textbf{\res{0}} &
\chg{It} premiered on amc in the us on october 31, 2010, but you can probably find \chg{it} on any basic cable channel like fox or hulu.\\
\addlinespace[0.08em]

\textbf{\res{4}} &
\chg{The show} premiered on amc in the us on October 31, 2010, but you can probably find \chg{The show} on any basic cable channel like Fox or hulu. (gate$(\res{4}) = 0.7619$)\\
\addlinespace[0.08em]

\textbf{\res{5}} &
\chg{It} premiered on \chg{AMC} in the \chg{US} on October 31, 2010, but you can probably find \chg{it} on any basic cable channel like Fox or \chg{Hulu}. (gate$(\res{5}) = 0.5368$)\\
\addlinespace[0.22em]

\midrule
\multicolumn{2}{@{}p{\dimexpr\columnwidth\relax}@{}}{
\texttt{Example 3 (ID: \detokenize{617___2--0})}
}\\
\\

$C$ &
were you aware that the famous musician Elvis' middle name was Aaron? no, i wasn't ! \ldots\ i don't know much about Elvis. Where is he from?\\
\addlinespace[0.15em]

\textbf{\res{0}} &
\chg{He} was born in Tupelo Mississippi, but relocated to Memphis when \chg{he} was 13.\\
\addlinespace[0.08em]

\textbf{\res{4}} &
\chg{Elvis} was born in Tupelo Mississippi, but relocated to Memphis when \chg{Elvis} was 13. (gate$(\res{4}) = 0.7891$)\\
\addlinespace[0.08em]

\textbf{\res{5}} &
\chg{He} was born in Tupelo, Mississippi, but relocated to Memphis when \chg{he} was thirteen. (gate$(\res{5}) = 0.5490$)\\

\bottomrule
\end{tabularx}
}

\caption{Representative DialFact instances used for qualitative analysis. Each block shows an excerpt of the dialogue context ($C$) and the resulting claim surface under the original claim (\res{0}), the scoped antecedent-substitution rewrite (\res{4}), and the decoder one-shot rewrite (\res{5}). Edited spans are marked with \chg{...}; BiCon-Gate scores for \res{4} and \res{5} are shown in parentheses.}
\label{tab:rewrites}

\endgroup
\end{table}

\begin{table*}[t]
\centering
\refstepcounter{subsection}
\subsection*{\thesubsection\ \ Model identifiers}

\footnotesize
\setlength{\tabcolsep}{6pt}
\renewcommand{\arraystretch}{1.12}

\begin{tabular}{@{}p{0.25\textwidth} p{\dimexpr0.75\textwidth-3\tabcolsep\relax}@{}}
\toprule
\textbf{Component} & \textbf{Model / identifier} \\
\midrule
Punctuation restoration (\res{2})
  & \texttt{oliverguhr/fullstop-punctuation-multilang-large} \cite{vandeghinste2023fullstoppunctuationsegmentationpredictiondutch} \\

True-casing (\res{3})
  & \texttt{bert-base-cased} \cite{devlin2019bertpretrainingdeepbidirectional} \\

Coreference resolver (\res{4})
  & Maverick coreference resolver \cite{martinelli-etal-2024-maverick} \\

Antecedent selector (\res{4})
  & \texttt{meta-llama/Llama-3.1-8B-Instruct} \cite{grattafiori2024llama3herdmodels} \\

Decoder rewrite (\res{5})
  & \texttt{Qwen/Qwen2.5-14B-Instruct} \cite{hui2024qwen2} \\

NLI verifier (FV-only) / BiCon-Gate (NLI scorer)
  & \texttt{MoritzLaurer/DeBERTa-v3-large-mnli-fever-anli-ling-wanli} \cite{laurer2024buildingefficientuniversalclassifiers} \\

BiCon-Gate embedding encoder
  & \texttt{intfloat/multilingual-e5-large} \cite{wang2024multilinguale5textembeddings} \\

Sparse retriever (IR)
  & BM25 (Pyserini) \\

Dense retriever (IR)
  & E5-large \cite{wang2024multilinguale5textembeddings} \\

Cross-encoder reranker (IR)
  & \texttt{BAAI/bge-reranker-large} \cite{chen2024bgem3embeddingmultilingualmultifunctionality} \\
\bottomrule
\end{tabular}

\caption{Model identifiers for third-party components used in our de-colloquialisation pipeline, retrieval, and verification experiments.}
\label{tab:model-identifiers}
\end{table*}

\begin{table*}[t]
  \centering
  \refstepcounter{subsection}
  \subsection*{\thesubsection\ \ Prompts for Decoder-Based Rewriting}
  \footnotesize
  \setlength{\tabcolsep}{6pt}
  \renewcommand{\arraystretch}{1.12}
  \begin{tabular}{@{}p{0.98\linewidth}@{}}
    \toprule
    \textbf{Prompt for decoder-based rewrite $R_5$} \\
    \midrule
    \textbf{System} \\
    Follow the instructions exactly. Do not add or change facts. \\[0.4em]
    \textbf{User} \\
    You are an expert editor who rewrites informal, chatty utterances into well-formed declarative English without changing their meaning. \\
    You will receive: (i) \textbf{Context}, a list of previous dialogue turns; and (ii) \textbf{Response}, the claim text to be normalised. \\
    Task: Rewrite \texttt{Response} into \texttt{New\_Response} by applying only the following operations. \\
    (1) Add missing sentence-ending punctuation and fix spacing around punctuation. \\
    (2) Fix capitalisation at sentence starts and for proper nouns. \\
    (3) Insert missing apostrophes (e.g., \textit{dont}$\rightarrow$\textit{don't},
        \textit{cant}$\rightarrow$\textit{can't},
        \textit{im}$\rightarrow$\textit{I'm}). \\
    (4) Expand all contractions to full forms (e.g., \textit{isn't}$\rightarrow$\textit{is not},
        \textit{aren't}$\rightarrow$\textit{are not},
        \textit{won't}$\rightarrow$\textit{will not},
        \textit{wouldn't}$\rightarrow$\textit{would not},
        \textit{I'm}$\rightarrow$\textit{I am},
        \textit{it's}$\rightarrow$\textit{it is},
        \textit{they're}$\rightarrow$\textit{they are},
        \textit{don't}$\rightarrow$\textit{do not},
        \textit{can't}$\rightarrow$\textit{cannot}). \\
    (5) If a pronoun in \texttt{Response} (\textit{this/that/it/he/she/they/these/those})
        has a unique, clear antecedent in \texttt{Context}, replace it with that antecedent
        phrase; if ambiguous, leave it unchanged. \\
    (6) Do not add, remove, or correct any facts, numbers, names, or dates; preserve
        the claim's semantics exactly. \\
    (7) Output only the rewritten text as one or more sentences, with no explanations,
        lists, or markdown. \\[0.4em]
    The input is formatted as:
    \texttt{Context (earliest$\rightarrow$latest): \{context\_lines\}},
    \texttt{Response: \{response\_text\}},
    \texttt{Output: (New\_Response only; no explanations)}. \\
    \bottomrule
  \end{tabular}
  \caption{Prompt used with \texttt{Qwen2.5-14B-Instruct} to generate the decoder-based one-shot rewrite ($R_5$). The prompt constrains the model to apply only surface normalisation and unambiguous pronoun substitution based on the provided context, without changing claim's meaning.}
  \label{tab:prompt}
\end{table*}

\begin{table*}[t]
  \centering
  \refstepcounter{subsection}
  \subsection*{\thesubsection\ \ Additional FV-only Results: Context Window Sensitivity}

  \footnotesize
  \setlength{\tabcolsep}{3.5pt}
  \renewcommand{\arraystretch}{1.08}

  \begin{adjustbox}{max width=\textwidth,center}
  \begin{tabular}{@{}c l *{6}{r}@{}}
  \toprule
  \multirow{2}{*}{\#Turns} & \multirow{2}{*}{Claim} & \multicolumn{6}{c}{FV-only (gold evidence)} \\
  \cmidrule(lr){3-8}
   &  & Acc & Macro-F1 & F1(S) & F1(R) & F1(NEI) & $\Delta$F1 \\
  \midrule

  \multirow{8}{*}{0}
   & $\res{0}$ & 65.31 & 63.56 & 46.47 & 80.37 & 63.84 & \multicolumn{1}{c}{--} \\
   & $\res{1}$ & 65.20 & 63.43 & 46.17 & 80.27 & 63.84 & -0.13 \\
   & $\res{2}$ & 65.27 & 63.53 & 46.50 & 80.26 & 63.84 & -0.03 \\
   & $\res{3}$ & 65.36 & 63.58 & 46.34 & 80.35 & 64.04 & +0.02 \\
   & $\res{4}$ & 65.02 & 63.41 & 47.38 & 79.38 & 63.48 & -0.15 \\
   \bestrow
   & \hspace{0.4em}+\textit{Gated} ($\res{4}$) & 65.90 & \best{64.83} & 51.15 & 79.06 & 64.28 & \best{+1.27} \\
   & $\res{5}$ & 60.84 & 59.50 & 44.91 & 70.82 & 62.76 & -4.06 \\
   & \hspace{0.4em}+\textit{Gated} ($\res{5}$) & 65.31 & 63.56 & 46.47 & 80.38 & 63.84 & +0.00 \\
  \midrule

  \multirow{8}{*}{2}
   & $\res{0}$ & 62.67 & 61.09 & 45.08 & 76.77 & 61.42 & \multicolumn{1}{c}{--} \\
   & $\res{1}$ & 62.66 & 61.08 & 45.00 & 76.78 & 61.46 & -0.01 \\
   & $\res{2}$ & 62.71 & 61.16 & 45.33 & 76.69 & 61.47 & +0.07 \\
   & $\res{3}$ & 62.80 & 61.22 & 45.25 & 76.75 & 61.67 & +0.13 \\
   & $\res{4}$ & 63.15 & 61.85 & 47.59 & 76.20 & 61.77 & +0.76 \\
   \bestrow
   & \hspace{0.4em}+\textit{Gated} ($\res{4}$) & 63.77 & \best{62.93} & 50.59 & 76.12 & 62.10 & \best{+1.84} \\
   & $\res{5}$ & 58.27 & 56.93 & 42.79 & 67.61 & 60.40 & -4.16 \\
   & \hspace{0.4em}+\textit{Gated} ($\res{5}$) & 62.23 & 60.40 & 42.92 & 77.00 & 61.28 & -0.69 \\
  \midrule

  \multirow{8}{*}{4}
   & $\res{0}$ & 62.90 & 61.73 & 49.67 & 74.51 & 61.12 & \multicolumn{1}{c}{--} \\
   & $\res{1}$ & 62.94 & 61.76 & 49.51 & 74.64 & 61.13 & +0.03 \\
   & $\res{2}$ & 62.89 & 61.72 & 49.58 & 74.55 & 61.03 & -0.01 \\
   & $\res{3}$ & 62.86 & 61.66 & 49.46 & 74.38 & 60.66 & -0.07 \\
   & $\res{4}$ & 63.37 & 62.40 & 51.63 & 74.13 & 61.45 & +0.67 \\
   \bestrow
   & \hspace{0.4em}+\textit{Gated} ($\res{4}$) & 63.91 & \best{63.29} & 54.07 & 73.98 & 61.82 & \best{+1.56} \\
   & $\res{5}$ & 58.54 & 57.61 & 46.81 & 65.84 & 60.19 & -4.12 \\
   & \hspace{0.4em}+\textit{Gated} ($\res{5}$) & 62.83 & 61.58 & 48.66 & 74.86 & 61.22 & -0.15 \\
  \midrule

  \multirow{8}{*}{6}
   & $\res{0}$ & 62.77 & 61.75 & 50.83 & 73.89 & 60.67 & \multicolumn{1}{c}{--} \\
   & $\res{1}$ & 62.82 & 61.80 & 50.79 & 73.93 & 60.67 & +0.05 \\
   & $\res{2}$ & 62.81 & 61.78 & 50.85 & 73.85 & 60.64 & +0.03 \\
   & $\res{3}$ & 62.73 & 61.72 & 50.91 & 73.69 & 60.57 & -0.03 \\
   & $\res{4}$ & 63.33 & 62.52 & 53.04 & 73.57 & 60.95 & +0.77 \\
   \bestrow
   & \hspace{0.4em}+\textit{Gated} ($\res{4}$) & 63.94 & \best{63.44} & 55.44 & 73.47 & 61.42 & \best{+1.69} \\
   & $\res{5}$ & 58.21 & 57.46 & 47.73 & 65.07 & 59.58 & -4.29 \\
   & \hspace{0.4em}+\textit{Gated} ($\res{5}$) & 62.88 & 61.77 & 49.90 & 74.32 & 61.07 & +0.02 \\
  \midrule

  \multirow{8}{*}{8}
   & $\res{0}$ & 62.69 & 61.71 & 50.98 & 73.87 & 60.56 & \multicolumn{1}{c}{--} \\
   & $\res{1}$ & 62.78 & 61.81 & 50.98 & 73.82 & 60.62 & +0.10 \\
   & $\res{2}$ & 62.72 & 61.73 & 50.95 & 73.67 & 60.59 & +0.02 \\
   & $\res{3}$ & 62.63 & 61.66 & 51.00 & 73.53 & 60.45 & -0.05 \\
   & $\res{4}$ & 63.21 & 62.43 & 53.16 & 73.32 & 60.82 & +0.72 \\
   \bestrow
   & \hspace{0.4em}+\textit{Gated} ($\res{4}$) & 63.64 & \best{63.18} & 55.40 & 73.23 & 60.91 & \best{+1.47} \\
   & $\res{5}$ & 58.13 & 57.41 & 47.95 & 64.86 & 59.42 & -4.30 \\
   & \hspace{0.4em}+\textit{Gated} ($\res{5}$) & 62.61 & 61.57 & 50.08 & 73.80 & 60.83 & -0.14 \\
  \bottomrule
  \end{tabular}
  \end{adjustbox}

  \caption{FV-only context-window sensitivity (gold evidence). Results for $k\in\{0,2,4,6,8\}$ context turns, reporting Accuracy and macro/class-wise F1 (\%) for each claim surface; $\Delta$F1 denotes the macro-F1 change relative to \res{0} at the same k.}
  \label{tab:fv-only-context}
\end{table*}

\clearpage
\begin{figure*}[t]
  \centering
  \refstepcounter{subsection}
  \subsection*{\thesubsection\ \ FV-only Macro-F1 $\Delta$ Heatmap}

  \includegraphics[
    width=\textwidth, height=0.25\textheight, keepaspectratio, trim=2mm 2mm 2mm 2mm,clip ]{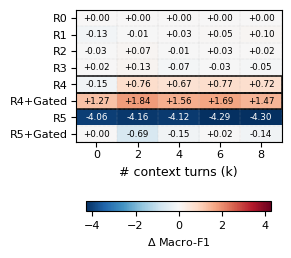}

  \caption{FV-only heatmap of $\Delta$macro-F1 across context turns $k\in\{0,2,4,6,8\}$, computed relative to \res{0} at the same $k$ (Table~\ref{tab:fv-only-context}). Boxed rows highlight \res{4} and \res{4}+\textit{Gated}, and the main setting is $k{=}2$.
  }
  \label{fig:heatmap-macro}
\end{figure*}

\begin{figure*}[t]
    \centering
    \refstepcounter{subsection}
    \subsection*{\thesubsection\ \ FV-only Class-wise $\Delta$F1 Heatmaps}

    \includegraphics[width=\linewidth,trim=2mm 2mm 2mm 2mm,clip]{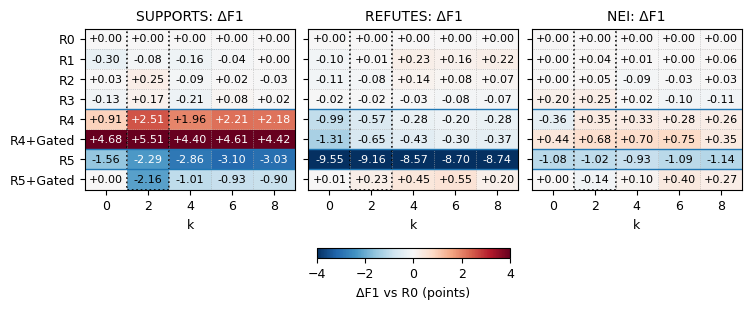}

    \caption{Class-wise FV-only $\Delta$F1 over context turns $k\in\{0,2,4,6,8\}$, computed from Table~\ref{tab:fv-only-context}.
    Colors are clipped for readability; annotated values show the true deltas, and the dotted box marks the main setting ($k{=}2$).}
    \label{fig:heatmap-classwise}
\end{figure*}

\end{document}